\pdfoutput=1

\documentclass[11pt]{article}
\usepackage[final]{acl}

\usepackage{times}
\usepackage{latexsym}
\usepackage[T1]{fontenc}
\usepackage[utf8]{inputenc}
\usepackage{microtype}
\usepackage{inconsolata}
\usepackage{graphicx}
\usepackage[table]{xcolor}
\usepackage{listings}
\usepackage{multirow}
\usepackage{pdfpages}
\usepackage{epsfig}
\usepackage{adjustbox}
\usepackage{amsmath}
\usepackage{amssymb}
\usepackage{comment}
\usepackage{caption}
\usepackage{subcaption}
\usepackage{textcomp}
\usepackage{relsize}
\usepackage{stmaryrd}
\usepackage{bbm}
\usepackage{rotating}
\usepackage{tabularray}
\usepackage{colortbl}
\usepackage{helvet}
\usepackage{courier}
\usepackage{natbib}
\usepackage{cleveref}
\usepackage{xspace}
\usepackage{enumitem}
\usepackage{makecell}
\usepackage{array}
\usepackage{threeparttable}
\usepackage[normalem]{ulem}
\usepackage{arydshln}
\usepackage{lipsum}
\usepackage{wrapfig}
\usepackage{soul}
\usepackage{booktabs}
\usepackage{pifont}
\usepackage{longtable}
\usepackage[utf8]{inputenc}
\usepackage{makecell} 
\definecolor{lightgray}{gray}{0.95}

\definecolor{green}{RGB}{36, 214, 36}
\definecolor{red}{RGB}{235, 30, 30}
\definecolor{lightredshade}{HTML}{dea9a9}
\definecolor{lightgreenshade}{HTML}{bce3bd}
\definecolor{lightblueshade}{HTML}{cacbe8}
\definecolor{MyDarkBlue}{rgb}{0,0.08,1}
\definecolor{MyDarkGreen}{rgb}{0.02,0.6,0.02}
\definecolor{MyDarkRed}{rgb}{0.8,0.02,0.02}
\definecolor{MyDarkOrange}{rgb}{0.40,0.2,0.02}
\definecolor{MyPurple}{RGB}{111,0,255}
\definecolor{MyRed}{rgb}{1.0,0.0,0.0}
\definecolor{MyGold}{rgb}{0.75,0.6,0.12}
\definecolor{MyDarkgray}{rgb}{0.66, 0.66, 0.66}

\definecolor{MyYellow}{rgb}{254, 246, 170}
\definecolor{MyBlue}{rgb}{170, 217, 251}
\definecolor{LuneBlue}{rgb}{0.11, 0.11, 0.43}
\newcommand{\colorDelta}[1]{%
  \ifnum#1>50\relax\cellcolor{green!65}+{#1}\%\else%
  \ifnum#1>40\relax\cellcolor{green!40}+{#1}\%\else%
  \ifnum#1>30\relax\cellcolor{green!30}+{#1}\%\else%
  \ifnum#1>20\relax\cellcolor{green!20}+{#1}\%\else%
  \ifnum#1>10\relax\cellcolor{green!10}+{#1}\%\else%
  \ifnum#1>0\relax\cellcolor{green!5}+{#1}\%\else%
  #1\%\fi\fi\fi\fi\fi\fi%
}

\crefname{section}{§}{§§}

\definecolor{fc}{HTML}{4677a2}
\definecolor{eg}{HTML}{7a4341}
\definecolor{sn}{HTML}{47734c}

\definecolor{boxblue}{HTML}{4677a2}
\definecolor{boxpink}{HTML}{7a4341}
\definecolor{boxgreen}{HTML}{47734c}

\definecolor{JungleGreen}{rgb}{0.16, 0.67, 0.53}
\definecolor{BitterSweet}{rgb}{1.0, 0.44, 0.37}
\definecolor{orangegray}{rgb}{1.0, 0.49, 0.0}
\definecolor{bluegray}{rgb}{0.4, 0.6, 0.8}
\newcommand{\colorC}[1]{%
  \pgfmathparse{#1>15 ? 1 : (#1>12 ? 2 : (#1>9 ? 3 : (#1>7 ? 4 : (#1>5 ? 4.5 : (#1>2 ? 6 : (#1>-1 ? 7 : (#1>-3 ? 8 : (#1>-6 ? 9 : (#1>- ? 10 : 11))))))))}%
  \ifcase\pgfmathresult\relax%
  \or \cellcolor{bluegray!90}{#1}%
  \or \cellcolor{bluegray!70}{#1}%
  \or \cellcolor{bluegray!50}{#1}%
  \or \cellcolor{bluegray!30}{#1}%
  \or \cellcolor{bluegray!15}{#1}%
  \or \cellcolor{bluegray!5}{#1}%
  \or \cellcolor{orangegray!5}{#1}%
  \or \cellcolor{orangegray!15}{#1}%
  \or \cellcolor{orangegray!30}{#1}%
  \or \cellcolor{orangegray!45}{#1}%
  \or \cellcolor{orangegray!60}{#1}%
  \else #1\fi%
}

\newcommand{\eg}{{\it e.g.},~}%
\newcommand{\ie}{{\it i.e.},~}%

\newcommand{\ourD}{\textsc{ToolHaystack}\xspace}

\newcommand{\xmark}{\textcolor{red}{\text{\sffamily X}}}
\newcommand{\gcheckmark}{\textcolor[HTML]{006400}{\checkmark}}

\definecolor{lightgray}{gray}{0.95}
\definecolor{highlightblue}{rgb}{0.85, 0.92, 1.0} 
\newcommand{\cmark}{\gcheckmark} 

\lstdefinestyle{python}{
    language=Python,
    basicstyle=\fontsize{8}{10}\ttfamily,
    keywordstyle=\color{blue},
    commentstyle=\color{gray},
    stringstyle=\color{black},
    showstringspaces=false,
    breaklines=true,
    breakindent=0pt,
    breakatwhitespace=false,
    escapeinside={(*@}{@*)}
}

\lstdefinestyle{cpp}{
    language=C++,
    basicstyle=\fontsize{8}{10}\ttfamily,
    keywordstyle=\color{blue},
    commentstyle=\color{gray},
    stringstyle=\color{green},
    showstringspaces=false,
    breaklines=true,
    breakindent=0pt,
    breakatwhitespace=false,
    escapeinside={(*@}{@*)}
}

\lstdefinestyle{plain}{
    basicstyle=\fontsize{8}{10}\ttfamily,
    keywordstyle=\color{blue},
    commentstyle=\color{gray},
    stringstyle=\color{green},
    showstringspaces=false,
    breaklines=true,
    breakatwhitespace=false,
    breakindent=0pt,
    escapeinside={(*@}{@*)}
}

\lstdefinestyle{python2}{
    language=Python,
    basicstyle=\fontsize{8}{10}\ttfamily,
    keywordstyle=\color{blue},
    commentstyle=\color{gray},
    stringstyle=\color{green},
    showstringspaces=false,
    breakatwhitespace=false,
    breaklines=true,
    breakindent=0pt,
    escapeinside={(*@}{@*)}
}

\lstdefinestyle{cpp2}{
    language=C++,
    basicstyle=\fontsize{8}{10}\ttfamily,
    keywordstyle=\color{blue},
    commentstyle=\color{gray},
    stringstyle=\color{green},
    showstringspaces=false,
    breaklines=true,
    breakindent=0pt,
    breakatwhitespace=false,
    escapeinside={(*@}{@*)}
}

\lstdefinestyle{sql}{
    language=SQL,
    basicstyle=\fontsize{8}{10}\ttfamily,
    keywordstyle=\color{blue},
    commentstyle=\color{green},
    stringstyle=\color{black},
    showstringspaces=false,
    breakatwhitespace=false,
    breaklines=true,
    breakindent=0pt,
    escapeinside={(*@}{@*)}
}

\lstdefinestyle{prompt}{
    language=Python,
    basicstyle=\fontsize{8}{10}\ttfamily,
    keywordstyle=\color{blue},
    commentstyle=\color{gray},
    stringstyle=\color{cppgreen},
    showstringspaces=false,
    breaklines=true,
    backgroundcolor=\color{bgcolor},
    keepspaces=true, 
    breakindent=0pt,
    breakatwhitespace=false,
    showspaces=false,   
    escapeinside={(*@}{@*)}
}
\lstdefinestyle{text}{
    basicstyle=\fontsize{8}{10}\ttfamily,
    showstringspaces=false,
    breaklines=true,
    backgroundcolor=\color{bgcolor},
    breakatwhitespace=false,
    breakindent=0pt,
    keepspaces=true,
    showspaces=false,   
    escapeinside={(*@}{@*)}
}

\usepackage[most]{tcolorbox}
\tcbuselibrary{listings}
\usepackage{xcolor}

\newtcblisting{tinycodebox}[2]{
  title=#1,
  width=#2,
  colframe=cyan!75!black,
  colback=green!5!white,
  listing only,
  listing options={
    basicstyle=\ttfamily\tiny,
    breaklines=true,
    breakatwhitespace=false,
    showstringspaces=false
  },
  boxrule=0.5mm,
  arc=1mm,
  fonttitle=\bfseries
}

\newtcblisting{smallcodebox}[2]{
  title=#1,
  width=#2,
  colframe=cyan!75!black,
  colback=green!5!white,
  listing only,
  listing options={
    basicstyle=\ttfamily\small,
    breaklines=true,
    breakatwhitespace=false,
    showstringspaces=false
  },
  boxrule=0.5mm,
  arc=1mm,
  fonttitle=\bfseries
}

\makeatletter
\renewcommand{\@fnsymbol}[1]{%
  \ifcase#1\or \dagger \else \@arabic{#1}\fi%
}
\makeatother

\title{\ourD: Stress-Testing Tool-Augmented Language Models \\in Realistic Long-Term Interactions}

\author{
\textbf{Beong-woo Kwak}\textsuperscript{1} \quad
\textbf{Minju Kim}\textsuperscript{1} \quad
\textbf{Dongha Lim}\textsuperscript{2} \quad
\textbf{Hyungjoo Chae}\textsuperscript{1} \\
\textbf{Dongjin Kang}\textsuperscript{\textbf{1}} \quad
\textbf{Sunghwan Kim}\textsuperscript{\textbf{1}} \quad
\textbf{Dongil Yang}\textsuperscript{\textbf{1}} \quad
\textbf{Jinyoung Yeo}\textsuperscript{\textbf{1}\dag} \\
\textsuperscript{1}Department of Artificial Intelligence, Yonsei University\\\textsuperscript{2}Department of Computer Science \& Engineering, Yonsei University \\
\texttt{\{bwoo.kwak,jinyeo\}@yonsei.ac.kr}
}

\begin{document}
\maketitle

\begin{abstract}
Large language models (LLMs) have demonstrated strong capabilities in using external tools to address user inquiries. However, most existing evaluations assume tool use in short contexts, offering limited insight into model behavior during realistic long-term interactions. To fill this gap, we introduce \ourD, a benchmark for testing the tool use capabilities in long-term interactions. Each test instance in \ourD includes multiple tasks execution contexts and realistic noise within a continuous conversation, enabling assessment of how well models maintain context and handle various disruptions. By applying this benchmark to 14 state-of-the-art LLMs, we find that while current models perform well in standard multi-turn settings, they often significantly struggle in \ourD, highlighting critical gaps in their long-term robustness not revealed by previous tool benchmarks.\footnote{Our code and data are available at \url{https://github.com/bwookwak/ToolHaystack} \\ \textsuperscript{\dag}Corresponding author.}
\end{abstract}

\section{Introduction}

Recent breakthroughs in Large Language Models (LLMs) have transformed their role toward LLM \textit{agents}; they can execute real-world tasks such as managing financial transactions and scheduling appointments  \citep{yao2022react,shinn2023reflexion}. Tool-augmented language model (TALM) benchmarks \citep{qin2023toolllm,li2023apibank,chen2024t} have played a critical role in evaluating tool-use capabilities based on simplified instruction-following setups where they autonomously execute external tools to address user goals. 

\begin{figure}
    \centering
    \small
    \includegraphics[width=1\columnwidth]{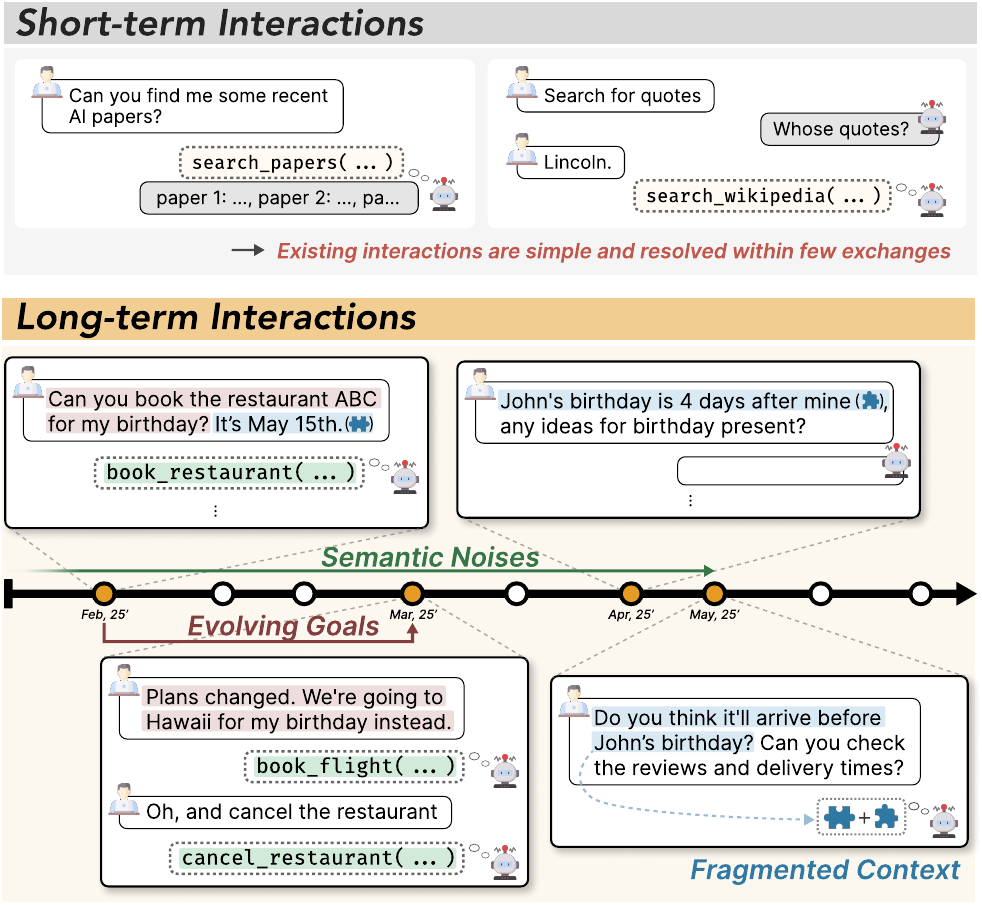}
    \caption{\textsc{ToolHaystack} addresses long-term interactions that include evolving goals, semantic noise, and fragmented context.}
    \label{fig:intro_figure}
\end{figure}

\begin{table*}[t!]
\small
\vspace{-6px}
\centering
\rowcolors{2}{white}{lightgray}
\begin{adjustbox}{width=0.9\textwidth}
\small
\begin{tabular}{lccccc}
\rowcolor{white}
\toprule
\textbf{Benchmarks} & 
\textbf{Multi-turn} & 
\textbf{Long Interaction} &
\textbf{Granular Complexity} &
\textbf{Distractor Tasks} &
\textbf{Avg. Turns} \\
\midrule
ToolBench             & \xmark & \xmark & \xmark & \xmark & 1\\
Fail-TaLMs            & \xmark & \xmark & \xmark & \xmark & 1\\
RoTBench              & \xmark & \xmark & \cmark & \xmark & 1\\
ComplexFuncBench      & \xmark & \xmark & \xmark & \xmark & 1\\
ToolDial              & \cmark & \xmark & \xmark & \xmark & 4.5\\
$\tau$-Bench          & \cmark & \xmark & \xmark & \xmark & --\\
MMTB                  & \cmark & \xmark & \xmark & \cmark & 5.3\\
HammerBench           & \cmark & \xmark & \cmark & \xmark & 2.2\\
BFCL-v3               & \cmark & \xmark & \xmark & \cmark & 4.2\\
\midrule
\textbf{\ourD}        & \textbf{\cmark} & \textbf{\cmark} & \textbf{\cmark} & \textbf{\cmark} & 32 \\
\bottomrule
\end{tabular}
\end{adjustbox}
\caption{Comparison of TALM benchmarks across key evaluation dimensions. \textbf{Long Interaction} captures extended conversations where long-term dependency plays a critical role; \textbf{Granular Complexity} denotes supporting fine-grained evaluation with varying difficulty levels; \textbf{Distractor Tasks} measure robustness under interleaved irrelevant tasks. The average number of turns $\tau$-Bench is marked as `--' as it does not use static dialogues for evaluation.}
\vspace{-4px}
\label{tab:dataset_comparison}
\end{table*}

Despite these advancements, the scope of most existing work is still limited short interaction, which fail to capture the behavior of LLM agents in \textbf{long-term interactions}. The queries of most existing benchmarks are simple (\eg \textit{``Book a restaurant named ....''}) which can be resolved within one or two exchanges. In contrast, many real-world tasks, such as travel planning or project coordination, cannot be resolved in short interactions \citep{wulongmemeval,maharana2024locomo} or steps \citep{hayati2025chain,press2023measuring}. For example, the user may want to perform tasks based on previous requests without restating all the details (\eg \textit{``Cancel the reservation you made for my birthday.''}) or request to collaborate with the agents across long time frame (\eg \textit{``Help me writing a research paper.''}). These tasks inherently require interactions that unfold and might evolve over time.

While recent efforts have extended test scenarios through multi-turn interactions \citep{li2023apibank} and API response simulations \citep{berkeley-function-calling-leaderboard,zhong2025complexfuncbench}, these approaches fall short in capturing the full complexity of long-term interactions. In real-world contexts, continuous exchanges between the user, the agent, and the external environment introduce various contextual noises—such as interleaved task flows \citep{castillo2024beyond}, redundant information \citep{bai2024longbench}, and shifting user goals \citep{yu2025multi}—that accumulate in the agent's context (Fig.\ref{fig:intro_figure}). The current lack of benchmarks that rigorously evaluate TALMs' robustness under such conditions limits their applicability to real-world scenarios, especially for long-term use cases like personal assistants.


In this paper, we aim to evaluate TALMs in realistic long-term interaction scenarios. To this end, we introduce \textbf{\ourD}, a realistic and composable test suite for TALMs. Specifically, we define long-term agent context as a highly noisy context where multiple user goals and partial task execution histories are entangled in a single context. Test instances are structured to simulate natural task flows by interleaving task-relevant (\textit{"needles"}) and irrelevant contexts (\textit{"haystack"})\footnote{We draw the analogy from the classic needle in a haystack problem \citep{kamradt2023needle} to emphasize the noisy nature of long-term interaction.}, posing unique challenges of identifying key information among distractions accumulated over time. Based on the structure, \ourD simulates six realistic scenarios derived from three core challenges (context recall, information shift, and missing context), each with two difficulty levels. 


We systematically evaluate 14 highly capable LLMs with tool-use capabilities, covering both open-source and proprietary models on \ourD. Our findings reveal that even highly capable LLMs with advanced long-context modeling \citep{bai2024longbench,an2024eval} still struggle to maintain consistency and robustness under long-term scenarios—suggesting that contemporary benchmarks may overestimate their readiness and reliability for real-world deployment. Beyond general performance, we also conduct detailed ablation studies to uncover what factors influence long-term success and where current models fail. 

Our main contributions are as follows:
\begin{itemize}[leftmargin=1em]
\itemsep0.05em 
\item We are the one of the pioneering works to explicitly formalize and highlight the critical challenge of long-term interaction robustness for TALMs.
\item We introduce \ourD, a benchmark to stress-test LLM agents in long-term scenarios.
\item We systematically evaluate 14 state-of-the-art LLM agents, showing key failure modes in long-term interactions and guidance for their robust real-world deployment.
\end{itemize}


\section{Related Work}
\paragraph{Realistic Evaluation of TALMs.}
To better approximate real-world conditions, benchmarks such as ToolDial \citep{shimtooldial}, HammerBench \citep{wang2024hammerbench}, and $\tau$-Bench \citep{yao2025taubench} introduce more naturalistic, multi-turn dialogues. Recent efforts like RoTBench \citep{ye2023rotbench} and FAIL-TALM \citep{trevino2025benchmarking} examine robustness under adversarial or erroneous conditions. However, these benchmarks remain largely constrained to single-session or short-horizon tasks, offering limited insight into TALM behavior over extended interactions. Some recent benchmarks such as MultiChallenge \citep{sirdeshmukh2025multichallenge} have attempted to evaluate language models in multi-turn scenarios systematically. Compared to MultiChallenge, ToolHaystack differs significantly in problem space, failure types, and evaluation design. MultiChallenge focuses on instruction-following in relatively short and clean conversations (~5 turns), with limited extent and control over noise structure. In contrast, ToolHaystack targets tool-use scenarios with an average of over 32 turns per instance, explicit argument chaining over long-horizon, and adjustable, scenario-specific noise injection. These settings better reflect real-world task complexity and allow us to uncover nuanced model brittleness under long-term, noisy interactions.

\begin{figure}[t!]
    \centering
    \small
    \includegraphics[width=\columnwidth]{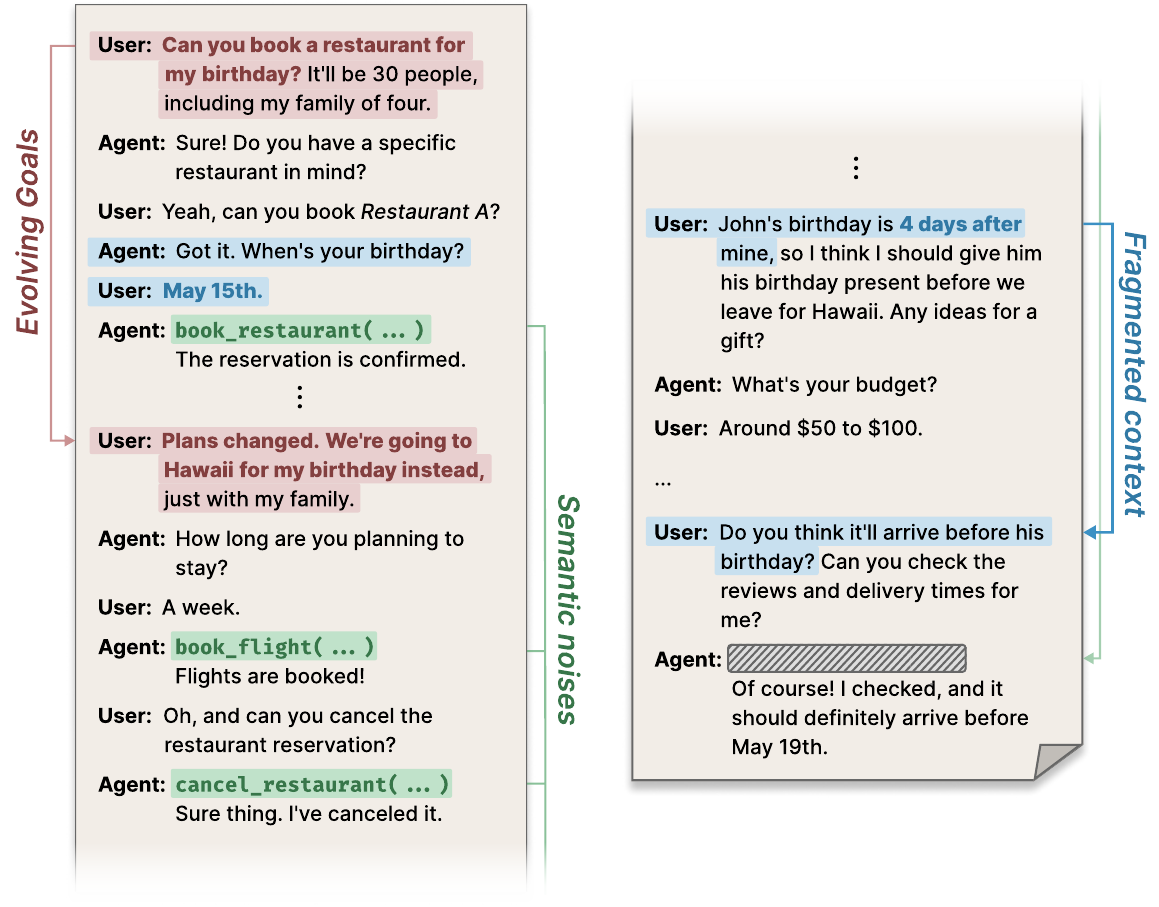}
    \caption{Real-world interactions between human and agent are intertwined, with natural contextual noise accumulated over time.}
    \label{fig:needles-in-haystack}
\end{figure}


\paragraph{Long-context Evaluation of TALMs.}
Recent benchmarks such as MMTB \citep{yu2025multi}, BFCL \citep{berkeley-function-calling-leaderboard}, and ComplexFuncBench \citep{zhong2025complexfuncbench}, ACEBench \citep{chen2025acebench}, APIGen-MT \citep{prabhakar2025apigen} have extended TALM evaluation into longer contexts and compositional tool use. However, they often rely on clean task structures with highly coherent subtasks (< 5) or simulate long context by concatenating tool outputs. In contrast, \ourD provides a long-context TALM benchmark with interleaved task flows and contextual noise. This exposes novel failure modes that are underexplored in prior evaluations. Tab.\ref{tab:dataset_comparison} compares \ourD with existing TALM benchmarks. Further discussions on related work are presented in Appendix~\ref{app:detailed_related_work}.



\section{Rethinking Evaluation of TALMs}
\begin{figure*}
    \centering
    \vspace{-6px}
    \small
    \includegraphics[width=0.99\linewidth]{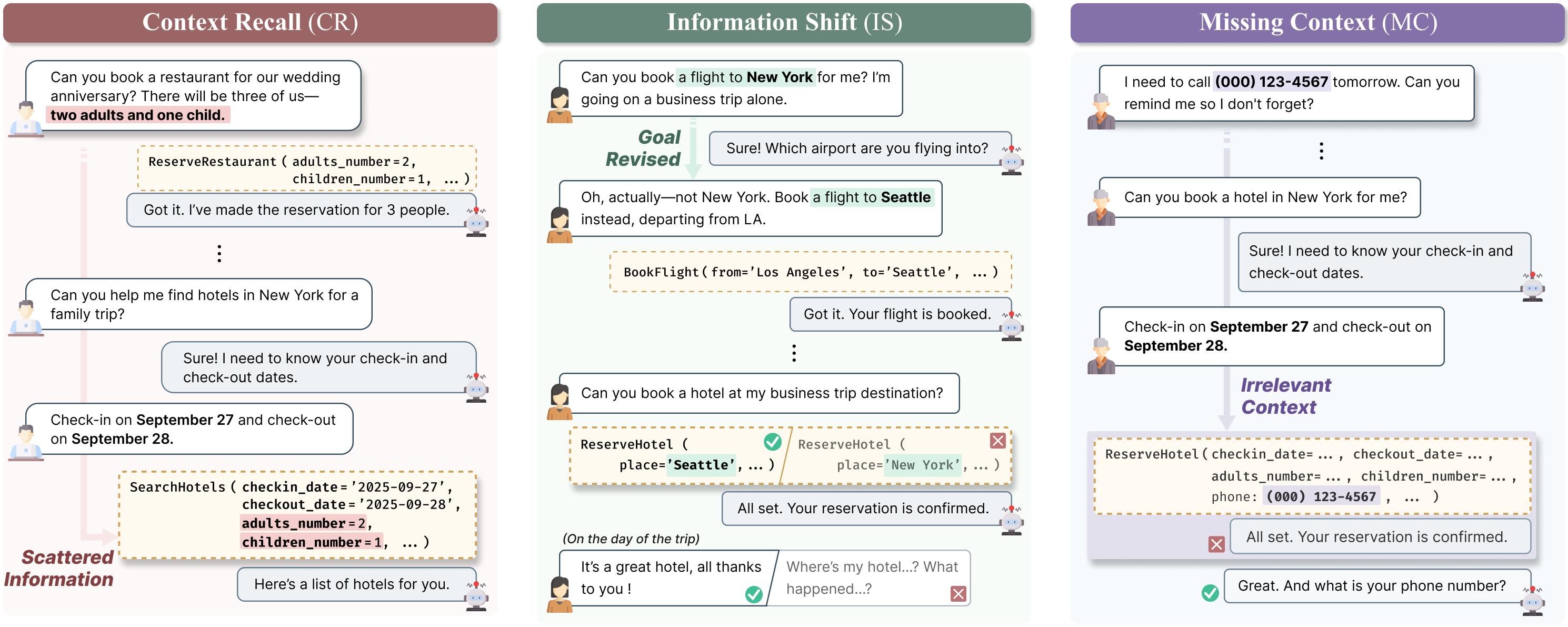}
    \caption{Illustration of the scenarios in \ourD (CR, IS, MC).}
    \vspace{-4px}
    \label{fig:scenarios}
\end{figure*}

Most LLM benchmarks are rapidly \textit{saturated} by frontier models \citep{sirdeshmukh2025multichallenge,phan2025humanity}. Likewise, previous TALM evaluation frameworks might have limited lifespans. This underscores a constant need for more realistic, challenging and adaptive evaluation frameworks that can keep pace with model advancements. 
We highlight that long-term interactions go beyond simple extension of standard multi-turn evaluation. 
\subsection{Long-Term \& Multi-Turn Tool-use} \label{sec:mt_and_lt}
The key distinction lies not just in the number of turns, it reflects the inherent \emph{complexity} of real-world tasks (Fig.\ref{fig:intro_figure}):
\begin{itemize}[leftmargin=1em]
\itemsep0.1em 
    \item \textbf{Fragmented context}: Relevant information is scattered across histories. The context often involves the execution history of multiple tasks.
    \item \textbf{Evolving goals}: User goals may shift over time. Models should adapt their response accordingly.
    \item \textbf{Semantic noise accumulation}: Irrelevant but plausible content such as tool responses from unrelated tasks can mislead models.
\end{itemize}

\subsection{Task Definition}
Let $\mathcal{T} = \{\text{tool}_1, \dots, \text{tool}_k\}$ be a set of tools, each taking a structured query as input and returning a result. A \textit{task} session consists of a sequence of interactions between the TALM, user, and environment. Let $S = \{s_1, \dots, s_T\}$ be the set of sessions, where each $s_t = \{x_t^1, a_t^1, r_t^1, \dots, x_t^n, a_t^n, r_t^n\}$ comprises user inputs $x_t^i$, model outputs $a_t^i$ (natural language response or tool calls), and tool response $r_t^i$ returned by the tool call (if applicable). Given the session history $H_t = \{s_1, \dots, s_{t-1}\}$ and the current session, the model chooses an action $a_t^i$ from the action space where a tool action is represented as a pair \textit{(tool\_name, arguments)}.

\section{\ourD: A Realistic and Composable Test Suite for Long-Term Tool Utilization}
We introduce \ourD, a benchmark to stress-test TALM's abilities in long-horizon tasks.
\subsection{Test Structure} \label{sec:dataset_structure}
The key principles of \ourD are: (1) The interaction should be aligned with how real-world users communicate with agents; and (2) The difficulty and contextual interference levels should be adjustable to keep the pace of model improvements. However, annotating such test is difficult and time-consuming even for human experts. 

To address this issue, we design a novel test structure for \ourD, inspired by Needle-In-A-Haystack (NIAH) dataset \citep{kamradt2023needle}. The key idea is to interleave different task sessions as a context instead of isolating task sessions (Fig.1). In \ourD, needles are defined as parts of the context that are essential to address the target task. The term haystack refers to the remaining (usually irrelevant) text, that acts as a distractor. This modularized structure enables us to control the difficulty of test instances by manipulating how we compose this haystack. For example, increasing the number of task sessions in the haystack can naturally increase the task difficulty.







\subsection{Test Scenarios} \label{sec:dataset_scenario}
To rigorously evaluate the capabilities of TALMs in long-term scenario, we construct a taxonomy of scenarios based on three essential abilities reflecting realistic long-term tool-use challenges: Context Recall, Information Shift, and Missing Context. Each scenario is further divided into two levels of difficulties (\ie \text{simple} and \text{complex}), yielding 6 scenarios in total. 



\subsubsection{Context Recall (CR)}
Users often refer back to previously mentioned details using pronouns or contextual ques. In this scenario, agents must ensure the continuity and relevance in dialogue by identifying and (if required) integrating relevant past information scattered across long conversations.
\begin{itemize}[leftmargin=1em]
\itemsep0.5em 
\item \textbf{CR-\textit{Single}:} A single relevant piece of information (the ``needle'') must be extracted from the haystack. This mirrors scenarios where a user refers to a specific past instruction or fact.
\item \textbf{CR-\textit{Multi}:} Multiple pieces of relevant information must be identified and integrated. This setting tests the model’s ability to consolidate information from various parts of the conversation, requiring both memory and reasoning.
\end{itemize}
\subsubsection{Information Shift (IS)}

Long-term user interactions are dynamic—users may revise goals, introduce new constraints, or provide updated information over time. This scenario evaluates whether the agent can track these changes, even when shifts are subtle or embedded within noise. Misalignment often leads to incorrect actions, like executing outdated plans.
\begin{itemize}[leftmargin=1em]
\itemsep0.5em 
\item \textbf{IS-\textit{Explicit}:} Changes are clearly stated by the user (\eg ``Actually, change the flight to May 10th instead of April 15th.''). Models must detect and adapt to these changes.
\item \textbf{IS-\textit{Implicit}:} Shifts are implied through indirect context cues (\eg ``I won't be able to leave that early anymore. Can you book something in the afternoon?''). This setting evaluates pragmatic understanding and contextual reasoning.
\end{itemize}

\subsubsection{Missing Context (MC)}
An often overlooked aspect of agent robustness is recognizing when essential information is absent (\eg missing API parameters) and abstaining from potentially erroneous responses to avoid failures.



\begin{itemize}[leftmargin=1em]
\itemsep0.5em 
\item \textbf{MC-\textit{Easy}:} Missing information is obvious, such as when no relevant entity or fact has ever been mentioned.
\item \textbf{MC-\textit{Hard}:} The gap is subtle or masked by irrelevant but similar-seeming information, requiring fine-grained discrimination between known and unknown details.
\end{itemize}

\begin{figure*}
    \centering
    \small
    \vspace{-6px}
    \includegraphics[width=\linewidth]{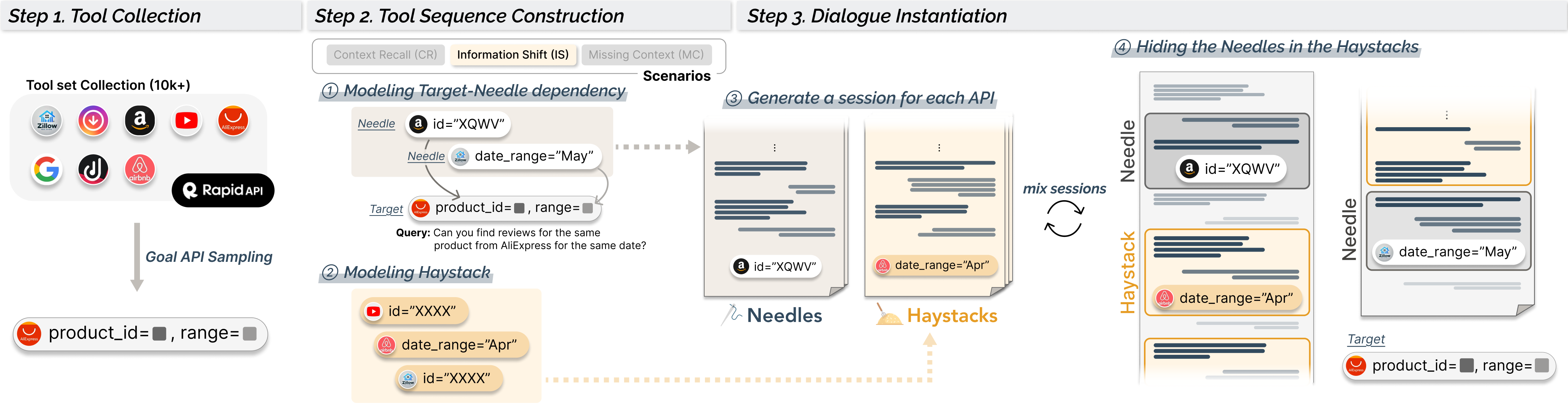}
    \caption{Overview of our three-stage dataset generation pipeline.}
    \vspace{-4px}
    \label{fig:data_pipeline}
\end{figure*}

\section{Building \ourD}
\label{sec:dataset_generation}
To ensure high-quality data while maintaining cost-efficiency, \ourD employs a multi-stage generation pipeline with three main phases: (1) Tool set collection; (2) Scenario-grounded tool sequence construction (3) Dialogue instantiation. Detailed process of generation process is presented in Appendix \ref{app:dataset_generation_detail}. 

\subsection{Tool Set Collection}
We begin by collecting over 13K real-world APIs from RapidAPI~\citep{rapidapi}, spanning 42 categories. Each API comes with a description of functionality and argument schema. 
We first sample a \textit{target API} to be invoked in the target session from the curated toolset.

\subsection{Scenario-Grounded Tool Sequence Construction}
To avoid incoherent sessions in test instances, we construct structured tool sequences as a blueprint for subsequent generation process. This approach enables us to generate test instances with coherent long-term dependencies between tools through shared arguments.
\paragraph{Modeling Target--Needle Dependencies.}

For a target API, we identify a set of \textit{needle} APIs that can form a long-term dependency with target API through shared arguments. For example, a hotel booking API may require a location and date, provided earlier by a flight search or calendar tool.
 
Each argument's dependency is identified using semantic matching (SentenceBERT \citep{reimers2019sentence}) and fine-grained LLM verification to ensure the semantic relevance and compatibility. We annotate each tool call with realistic argument values using LLM prompting. Specifically, a tuple of 4 elements is given to the model \ie target API, needle APIs, a scenario type (\eg CR, IS, and MC), situational context. Here, situation context is introduced to enhance the authenticity and consistency of the argument values. We pre-constructed a set of situational contexts for each goal APIs based on the characteristics and requirements of each target API and randomly use one of them as a situational context. 

\paragraph{Modeling the Haystack.}
We configure a set of tools to be used in \textit{haystack} (distractor) sessions based on the goal API and the selected scenario (\eg CR, IS, and MC). An LLM is employed to compose the APIs to be used for distractors and their arguments, guided by each scenario. For instance, in MC-Easy, we use semantically disjoint tools, while in MC-Hard we include distractors that mimic needle semantics. 

The prompt for second stage filtering of modeling target-needle dependencies stage and scenario grounding are presented in Appendix. We carefully craft prompts for applying each test scenario (CR, IS, MC) to build target-needle dependencies and model the haystack.

\subsection{Dialogue Instantiation}
With structured tool sequences fully annotated, we convert them into naturalistic dialogues. This involves rendering tool arguments into plausible user utterances and assistant responses. The needle and haystack interactions are interleaved. This results in dialogues where key information is embedded in distracting context. All instances undergo automated validation to ensure API correctness, argument consistency, and logical coherence. Then, human reviewers inspect each dialogue for naturalness and functional integrity. Only instances passing both validation stages are retained in \ourD, ensuring high-quality data.

\subsection{Quality Assurance}
We employ multiple frontier LLMs (GPT-4o, Claude 3.5, Gemini 2.5, LLaMA 3.3) during generation to reduce model-specific artifacts. Automated validation enforces constraints such as correct argument use, proper tool syntax, and scenario fidelity. A subset of generated dialogues undergoes a filtering process, where we prompt gpt-4o to strictly evaluate if the dialogue violates the rules for the target category. Moreover, to ensure the quality of \ourD, we employed GPT-4o as a judge to automatically filter out dialogues that do not satisfy our scenario definitions. To validate the effectiveness of this automated filtering process, we sampled 60 dialogues (10 from each category) from the unfiltered dataset and compared the results against human annotations. The evaluation yielded a Matthews Correlation Coefficient (MCC) of 0.884, indicating a strong agreement between the automated and human filtering outcomes.

\section{Experiments} \
\subsection{Experimental Setup}
\paragraph{Models.}
We evaluate a comprehensive set of highly capable TALMs. \textbf{Closed-source models} include GPT family (4o, 4.5-Preview, and 4o-mini), Claude family (3.7-Sonnet, 3-Opus), Gemini Family (2.5-Pro, 2.0-Flash-001), Amazon-Nova-Pro, Mistral-large and Grok-3. \textbf{Open-source models} are Qwen2.5-72B-Instruct, Llama-3.3-70B-Instruct, DeepSeek-V3, ToolACE-2-Llama-3.1-8B, watt-tool-8B and xLAM-2-32b-fc-r. Note that ToolACE-2-Llama-3.1-8B, watt-tool-8B and xLAM-2-32b-fc-r are finetuned for function calling tasks. Models are selected to cover a diverse and competitive set of top-performing TALMs, particularly those demonstrating strong capabilities in recent multi-turn tool benchmarks (\eg BFCL).
\paragraph{Retrieval} As providing all the API documents is infeasible, a subset of feasible APIs is typically given to TALM through semantic matching. In \ourD, the model is tested in oracle setup with the 5 API documents, where a ground-truth API document and 4 APIs are given to the model based on their semantic relevance to the target session.

\begin{table*}[h!]
    \centering
    \renewcommand{\arraystretch}{1.0}
    \setlength{\tabcolsep}{4pt}
    \resizebox{0.98\textwidth}{!}{
    \begin{tabular}{l ccr ccr ccr cc}
        \toprule
        \multirow{2}{*}{\textbf{Method}} 
        & \multicolumn{3}{c}{\textbf{Context Recall (\%)}} 
        & \multicolumn{3}{c}{\textbf{Information Shift (\%)}} 
        & \multicolumn{3}{c}{\textbf{Missing Context (\%)}} 
        & \multirow{2}{*}{\textbf{Base}}
        & \multirow{2}{*}{\textbf{Avg.}} \\
        \cmidrule(lr){2-4} \cmidrule(lr){5-7} \cmidrule(lr){8-10}
        & \textit{Single} & \textit{Multi} & $\Delta$ ~~~~
        & \textit{Explicit} & \textit{Implicit} & $\Delta$ ~~~~
        & \textit{Easy} & \textit{Hard} & $\Delta$ ~~~~
        & & \\
        \midrule
        \rowcolor{gray!20}
        \multicolumn{12}{l}{\textbf{\textit{Closed-source models}}} \\
        GPT-4o & 63.44 & 50.00 & +13.44 & 31.67 & 28.57 & +3.10 & 21.52 & 20.00 & +1.52 & \textbf{83.33} & 35.87 \\
        GPT-4.5-Preview & \textbf{70.97} & 53.85 & +17.12 & 38.33 & \textbf{35.71} & +2.62 & \textbf{29.11} & 24.71 & +4.40 & 72.22 & \textbf{42.11} \\
        GPT-4o-mini & 56.99 & 38.46 & +18.53 & 38.33 & 27.38 & +10.95 & \textbf{29.11} & 12.94 & +16.17 & 74.07 & 33.87 \\
        Claude-3.7-Sonnet & 52.69 & 55.77 & -3.08 & 25.00 & 21.43 & +3.57 & 11.39 & 8.24 & +3.15 & 64.81 & 29.09 \\
        Claude-3-Opus & 34.41 & 30.77 & +3.64 & 8.33 & 9.52 & -1.19 & 5.06 & 4.71 & +0.35 & 79.63 & 15.47 \\
        Gemini-2.5-Pro & 48.39 & 57.69 & -9.30 & 20.00 & 25.00 & -5.00 & 12.66 & 12.94 & -0.28 & 59.26 & 29.45 \\
        Amazon-Nova-Pro-v1.0 & 54.84 & 44.23 & +10.61 & 35.00 & 23.81 & +11.19 & 27.85 & \textbf{31.76} & -3.91 & 52.78 & 36.25 \\
        Mistral-large-2407 & 59.14 & 54.08 & +5.06 & 26.67 & 25.51 & +1.16 & 20.32 & 23.00 & -2.68 & 57.41 & 34.79 \\
        Grok-3-beta & 64.52 & 61.54 & +2.98 & 36.67 & \textbf{35.71} & +0.96 & 15.19 & 20.00 & -4.81 & 66.67 & 38.94 \\
        \midrule
        \rowcolor{gray!20}
        \multicolumn{12}{l}{\textbf{\textit{Open-source models}}} \\
        Qwen2.5-72B-Instruct & 59.14 & \textbf{65.38} & -6.24 & 31.67 & 28.57 & +3.10 & 16.46 & 5.88 & +10.58 & 62.96 & 34.52 \\
        Llama-3.3-70B-Instruct & 45.16 & 53.85 & -8.69 & 20.00 & 20.24 & -0.24 & 0.00 & 0.00 & +0.00 & 74.07 & 23.21 \\
        DeepSeek-V3 & 55.91 & 59.62 & -3.71 & \textbf{40.00} & 28.57 & +11.43 & 27.85 & 18.82 & +9.03 & 68.52 & 38.46 \\
        ToolACE-2-Llama-3.1-8B & 23.66 & 17.31 & +6.35 & 10.00 & 4.67 & +5.33 & 0.00 & 0.00 & +0.00 & 70.37 & 9.27 \\
        watt-tool-8B & 15.05 & 5.77 & +9.28 & 8.33 & 2.38 & +5.95 & 0.00 & 0.00 & +0.00 & 62.96 & 5.26 \\
        xLAM-2-32b-fc-r & 50.62 & 51.06 & -0.44 & 26.92 & 25.33 & +1.59 & 0.00 & 0.00 & +0.00 & 69.81 & 25.66 \\
        \bottomrule
    \end{tabular}
    }
    \caption{Main results. $\Delta$ denotes the gap between simple and complex levels. Base denotes BFCL score; Avg. is the mean performance across metrics.}
    \label{tab:main_table}
\end{table*}

\paragraph{Evaluation Metric.} We use call accuracy which calculates the proportion of correct function calls:
\begin{equation}
    \small
    \text{Call Acc} = \frac{\sum_{i=1}^N c_i}{\sum_{i=1}^N n_i}
\end{equation}
where $N$ is the number of samples, $n_i$ is the total number of function calls in $i$-th sample, and $c_i$ is the number of correct function calls in sample $i$.
\paragraph{Base Test Set (Multi-turn).} 
To provide deeper insight, we generate a multi-turn counterpart of our dataset generated by standard self-instruct approach previously used in the literature \citep{qin2023toolllm}. We use the same goal APIs for generating short multi-turn tool-use instances. All necessary information to complete the task is revealed within a few turns, without any distractions.

\subsection{Main Results} \label{sec:exp_main}

We present the \ourD results in Table~\ref{tab:main_table}, evaluating 17 TALMs across six long-term interaction scenarios. 
\paragraph{Overall Performance.} Among all models, \texttt{GPT-4.5-Preview} achieved the highest average score (42.11\%), followed by \texttt{Grok-3-beta} (38.94\%) and \texttt{DeepSeek-V3} (38.46\%). While current TALMs show competitive performance in multi-turn settings (\ie Base), their accuracy significantly deteriorates in \ourD. For instance, \texttt{GPT-4o} drops from 83.33\% in Base to 35.87\%, demonstrating the complexity of long-term tool-use.

\paragraph{Noise Robustness.}
The performance gap between simple and complex sub-scenarios shows the robustness of models against contextual noises. We observe GPT family models struggles to retain their performance in CR-Single and CR-Multi. For example, \texttt{GPT-4.5-Preview}, one of the strongest model in \ourD, underperform significantly in CR-Multi. In contrast, Claude family models and \texttt{Grok-3-beta} have experienced marginal performance drops by contextual noises, showing under 5\% of degradation across scenarios consistently. 


\paragraph{Closed- vs. Open-source Models.}
Closed-source models consistently outperform open-source models under noisy settings. While general-purpose open-source models (\texttt{Qwen2.5}, \texttt{DeepSeek-V3}) demonstrate more stable performance across scenarios, they still lag behind top proprietary models. Notably, open-source finetuned models like \texttt{ToolACE} and \texttt{watt-tool} achieve generally achieve poor accuracy of less than 10\%, despite their high BFCL scores. We further examine this by extending the comparison between multi-turn and long-term tool-use performance in the following section.

\subsection{Limitation of Multi-turn Benchmarks for Evaluating Long-term Tool Use}
\label{sec:exp_mt_and_lt}
\begin{figure}[t!]
    \centering
    \vspace{-5pt}\raggedright
    \includegraphics[width=\linewidth]{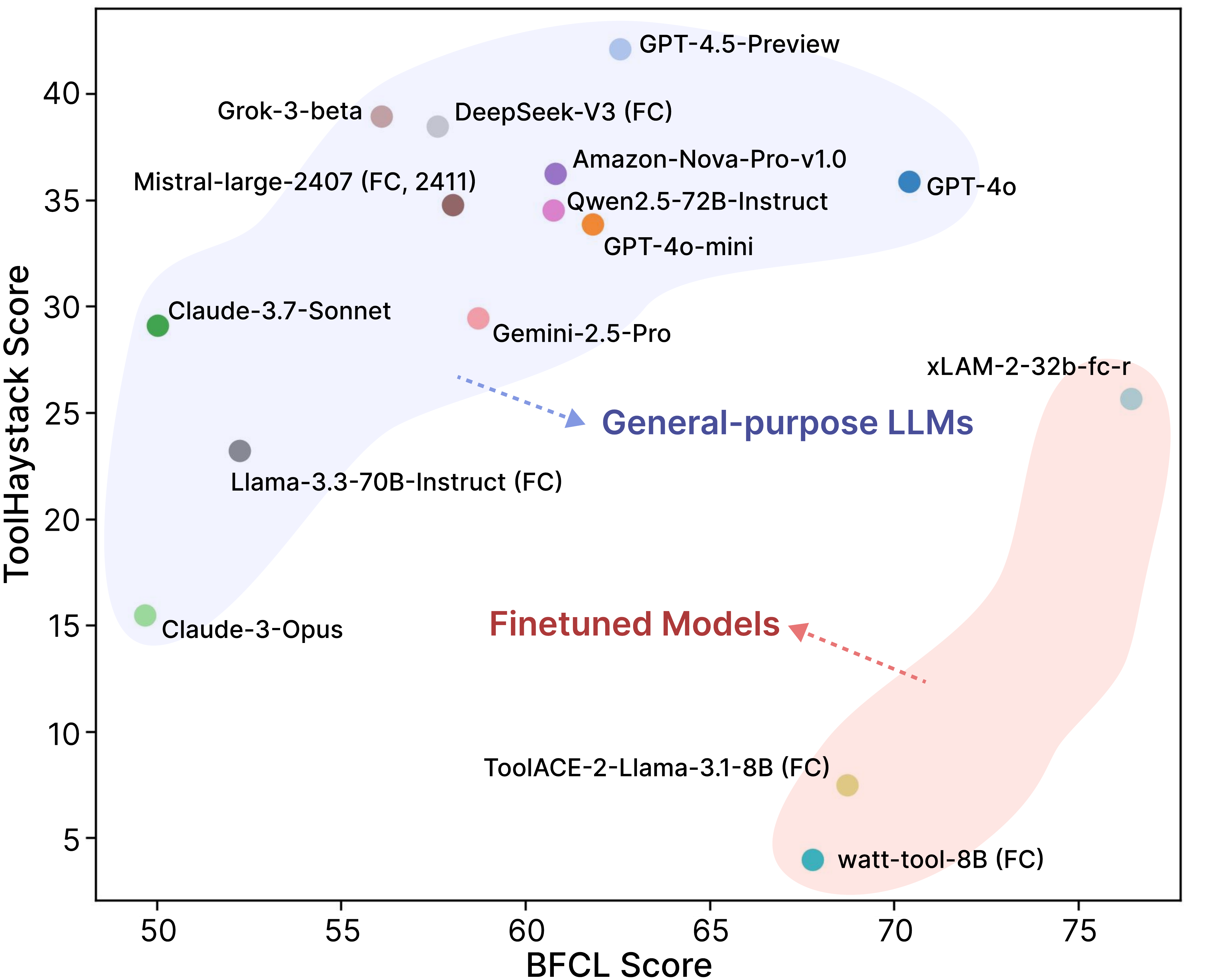}
    \caption{Performance comparison between BFCL (multi-turn) and \ourD (long-term) score.}
    \label{fig:bfcl_vs_ours}
\end{figure}

To investigate whether existing multi-turn benchmarks can reliably reflect long-term tool use performance, we compare performance on BFCL (multi-turn) and \ourD (long-term). As shown in Figure~\ref{fig:bfcl_vs_ours}, a positive correlation between the two benchmarks is observed within each group (general-purpose LLMs and finetuned models).
However, while considering all models, we find a notable discrepancy: finetuned models achieve high scores on BFCL but perform poorly on \ourD.
This gap suggests that multi-turn benchmarks may fail to capture the complexity and robustness required for long-term interactions. One possible explanation is an out-of-distribution (OOD) problem---models trained on conventional tool-learning corpus struggle to generalize to diverse, noisy and unpredictable nature of real-world scenarios.
As fine-tuning TALM is considered a promising direction to build personal agents, the lack of generalization in long-term tool-use can be a critical issue, highlighting the importance of evaluating tool use capabilities of LLMs during long-term interaction such as \ourD.

\begin{table}[t!]
    \centering
    \begin{adjustbox}{width=\columnwidth}
        \begin{tabular}{lcccccc}
        \toprule
        \textbf{Distance} & \textbf{0} (recent) & \textbf{1} & \textbf{2} & \textbf{3} & \textbf{4} & \textbf{5} (past)\\
        \midrule
        Closed    & 94.44 & 41.67 & 61.11 & 75.00 & 44.44 & 55.56 \\
        General & 83.33 & 25.00 & 50.00 & 66.67 & 33.33 & 58.33 \\
        TALM    & 33.33 & 0.00 & 33.33 & 41.67 & 25.00 & 50.00 \\
        \bottomrule
        \end{tabular}
    \end{adjustbox}
    \caption{Accuracy by evidence position (distance from evidence) across closed-source models, general-purpose open-sourced models, and finetuned TALMs. The numbers are averaged scores of each model group.}
    \vspace{-5pt}
    \label{tab:accuracy_by_evidence_position}
\end{table}

\subsection{Controlled Analysis on Long-Context Utilization of TALMs}\label{sec:controlled}
To isolate the impact of long-context challenges in tool-augmented settings, we conduct a controlled ablation study by systematically modifying the contextual structure in \ourD.
\paragraph{Positional Bias in TALMs.}
We analyze how model performance varies with the position of the key evidence ("needle") in the context. We report the averaged accuracy of models of three groups of models, namely Closed-source models, Open-source general purpose models, and finetuned TALMs, as a function of the number of sessions between the session containing the relevant evidence and the target session. As shown in Table~\ref{tab:accuracy_by_evidence_position}, closed-source models exhibit a strong recency bias, achieving peak performance (94.44\%) when the evidence is immediately adjacent to the query (distance 0). General-purpose open-source models follow a similar pattern but with lower overall robustness. In contrast, finetuned TALMs struggle across all positions. Overall, the findings underscore that positional biases significantly affect tool invocation accuracy.


\begin{figure}
  \centering
  \vspace{-5px}
  \includegraphics[width=0.8\columnwidth]{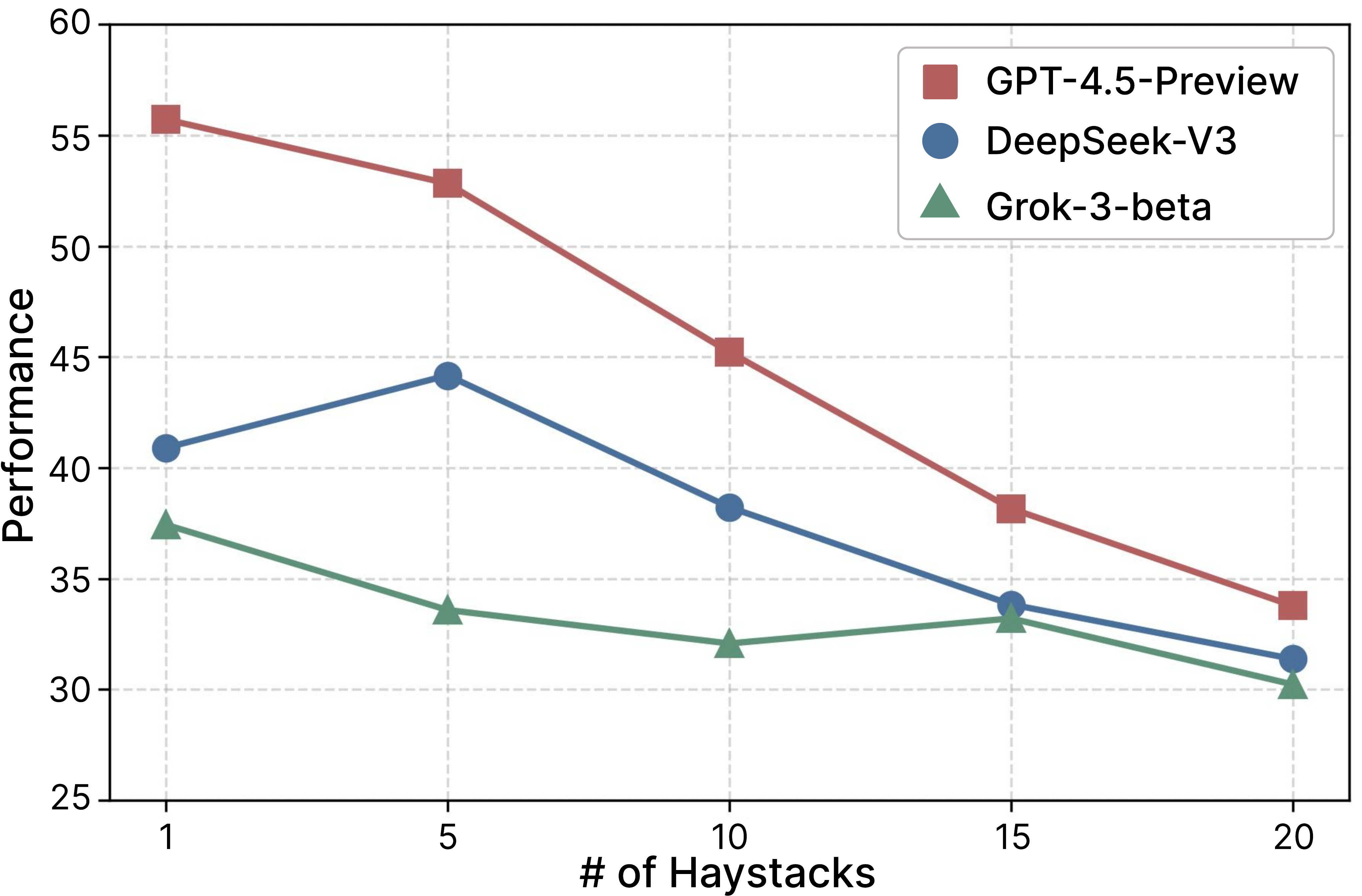}
  \caption{
  Performance degradation occurs as the number of distractor haystacks increases.
  }
  \label{fig:num_haystack}
\end{figure}

\paragraph{Impact of Distractors.} 
We investigate how increasing the number of haystack sessions lead to decreased performance. We select top-performing TALMs in \ourD and assess them on the samples with varying number of haystacks. In Figure~\ref{fig:num_haystack}, model performance consistently declines as the number of haystacks increases. For example, \texttt{GPT-4.5-Preview}'s accuracy drops sharply from 56\% to 32\% when the number of distractors grows from 1 to 20. Interestingly, the performance gap between \texttt{GPT-4.5} and other models narrows at higher noise levels, suggesting a saturation point beyond which all models are overwhelmed by irrelevant context. 

{\renewcommand{\arraystretch}{1.3}
\begin{table}[t!]
\centering
\footnotesize  
\setlength{\tabcolsep}{4pt}  
\begin{tabular}{p{0.95\linewidth}}  
\toprule
\textbf{Sample Dialogue from \textit{Missing Context-Hard}}: \\
(Previous sessions) \\
...\\
\textit{User}: Hey, I need to check the domains hosted on the same IP as a suspicious domain. Use the domain I mentioned before, and let's go with the page number we talked about last time. \\
\midrule
\textbf{Ground-truth Response}: \\
\begin{minipage}{\linewidth}
\footnotesize  
\begin{verbatim}
{
  "api_name": "Search domains",
  "parameters": {
    "q": "suspiciousdomain.com",
    "type": "WEBIP",
    "pagenum": "MISSING"
  }
}
\end{verbatim}
\end{minipage} \\
\midrule
\textbf{GPT-4o-mini Prediction}: \\
\begin{minipage}{\linewidth}
\footnotesize  
\begin{verbatim}
{
  "api_name": "Search domains",
  "parameters": {
    "q": "suspiciousdomain.com",
    "type": "WEBIP",
    "pagenum": "3"
  }
}
\end{verbatim}
\end{minipage} \\
\midrule
\textbf{DeepSeek-V3 Output}: \\
\begin{minipage}{\linewidth}
\footnotesize  
\begin{verbatim}
{
  "api_name": "Search domains",
  "parameters": {
    "q": "suspiciousdomain.com",
    "type": "WEBIP",
    "pagenum": "1"
  }
}
\end{verbatim}
\end{minipage} \\
\bottomrule
\end{tabular}
\caption{
An example from the \textit{Missing Context-Hard} setting showing the dialogue, ground-truth response, and model predictions. GPT-4-o-mini predicted pagenum as "3" which is a random number and DeepSeek-V3 predicted pagenum as "1" which is a default value.
}
\label{tab:case_study}
\end{table}
}

\subsection{Error Analysis through the Lens of Hallucination}
\begin{figure}
    \centering
    \includegraphics[width=0.99\linewidth]{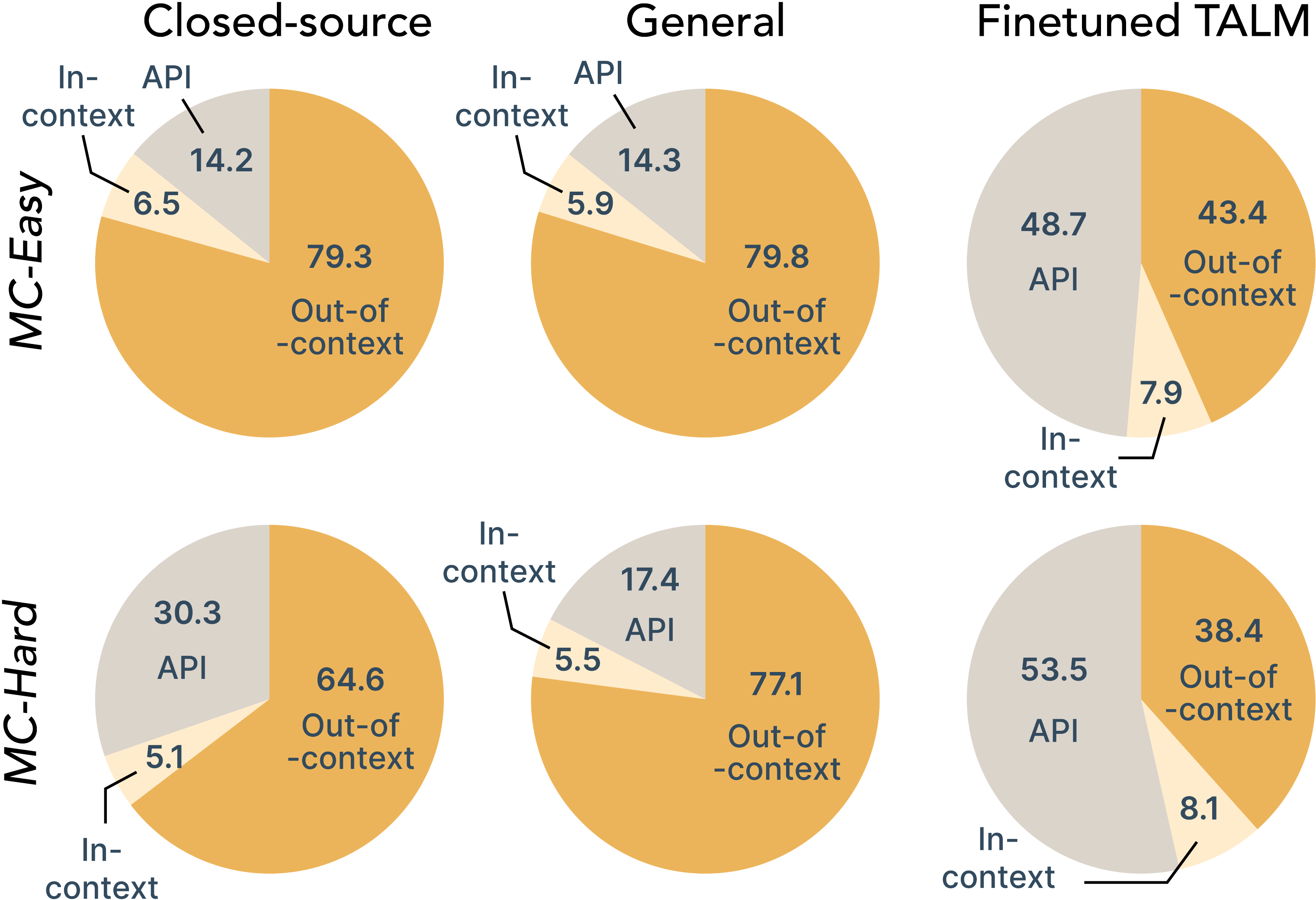}
    \caption{Compact error breakdown by model and context shown as percentages (API: wrong API name, In/Out-of-context: wrong parameter values filled with entities appearing in/out of context).}
    \label{fig:compact_error_distribution}
\end{figure}

As interaction history accumulates over time, TALMs are increasingly prone to hallucinating arguments or selecting incorrect tools by referring to irrelevant yet semantically-related context. We investigate model failures in the Missing Context (MC) scenarios where current TALMs struggle the most by analyzing the distribution of error types across model classes. Figure~\ref{fig:compact_error_distribution} presents the relative proportions of three key error types: wrong API selection, hallucinated arguments based on misleading in-context cues (In-context), and hallucinations derived from entirely absent information (Out-of-context). The numbers are averaged within each model group. 

\paragraph{Error Trends.} Closed and General-purpose models exhibit highly skewed error distributions, with over 75\% of their failures attributable to Out-of-context hallucinations in MC-Easy and similarly high proportions in MC-Hard (63\% and 77\%). We find that several models tend to copy default values of missing parameter or filling out randomly generated strings without proper verification of contextual sufficiency. Finetuned TALMs, in contrast, show a different error distribution. In MC-Easy, TALMs reduce Out-of-context hallucinations to 43.39\% and In-context hallucinations remain low (7.94\%), while API errors account for nearly half (48.68\%) of the total errors, suggesting that TALMs are more cautious about fabricating arguments outright from missing context (lower Out-of-context) but often misidentify the appropriate tool (API error), particularly under harder conditions. See Table~\ref{tab:case_study} for detailed case studies illustrating model behaviors.

\subsection{Does CoT Prompting Elicit Better Performance?}
We investigate whether CoT also improves robustness in long-term tool-use settings, where contextual entanglement, evolving goals, and noisy discourse severely challenge model reliability.

\paragraph{Setup.}
We conduct a controlled comparison of five models showing competitive performance in our dataset—\texttt{GPT-4o}, \texttt{GPT-4.5-Preview}, \texttt{Amazon-Nova-Pro-v1.0}, \texttt{Grok-3-beta}, and \texttt{DeepSeek-V3} across the test scenarios in \ourD. 
\begin{table}[t!]
    \centering
    \small
    \renewcommand{\arraystretch}{1.0}
    \setlength{\tabcolsep}{4pt}
    \resizebox{\columnwidth}{!}{
    \begin{tabular}{l cc cc cc c}
        \toprule
        \multirow{2}{*}{\textbf{Method}} 
        & \multicolumn{2}{c}{\textbf{CR (\%)}} 
        & \multicolumn{2}{c}{\textbf{IS (\%)}} 
        & \multicolumn{2}{c}{\textbf{MC (\%)}} 
        & \multirow{2}{*}{\textbf{Avg.}} \\
        \cmidrule(lr){2-3} \cmidrule(lr){4-5} \cmidrule(lr){6-7}
        & \textit{S} & \textit{M}
        & \textit{E} & \textit{I}
        & \textit{E} & \textit{H}
        & \\
        \midrule
        \textbf{GPT-4o} & \textbf{63.44} & 50.00 & \textbf{31.67} & 28.57 & \textbf{21.52} & \textbf{20.00} & 35.87 \\
        \quad+ CoT & 58.06 & \textbf{69.23} & 30.00 & \textbf{30.95} & 18.99 & 16.47 & \textbf{37.95} \\
        \midrule
        \textbf{GPT-4.5-Preview} & \textbf{70.97} & 53.85 & \textbf{38.33} & 35.71 & 29.11 & \textbf{24.71} & 42.11 \\
        \quad+ CoT & \textbf{70.97} & \textbf{61.54} & 30.00 & \textbf{42.86} & \textbf{31.65} & 22.35 & \textbf{43.23} \\
        \midrule
        \textbf{Amazon-Nova} & \textbf{54.84} & 44.23 & \textbf{35.00} & 23.81 & \textbf{27.85} & \textbf{31.76} & \textbf{36.25} \\
        \quad+ CoT & 52.69 & \textbf{51.92} & 25.00 & \textbf{26.19} & 17.72 & 25.88 & 33.23 \\
        \midrule
        \textbf{Grok-3-beta} & \textbf{64.52} & 61.54 & 36.67 & 35.71 & \textbf{15.19} & 20.00 & 38.94 \\
        \quad+ CoT & 59.14 & \textbf{71.15} & \textbf{38.33} & \textbf{36.90} & 13.92 & \textbf{23.53} & \textbf{40.83} \\
        \midrule
        \textbf{DeepSeek-V3} &\textbf{55.91} & 59.62 & \textbf{40.00} & 28.57 & \textbf{27.85} & \textbf{18.82} & \textbf{38.46} \\
        \quad+ CoT & 51.61 & \textbf{71.15} & 36.67 & \textbf{33.33} & 17.72 & 15.29 & 37.63 \\
        \bottomrule
    \end{tabular}
    }
    \caption{The results of each method are evaluated using the same setup as the main results. Rows listing only
the model name correspond to the vanilla (zero-shot)
setting. The detailed prompts for CoT are in the Appendix.
Note that Amazon-Nova stands for \texttt{Amazon-Nova-pro-v1}.
}
    \label{tab:cot_result}
\end{table}

\paragraph{Results.}
Our findings, summarized in Table~\ref{tab:cot_result}, reveal mixed effects of CoT prompting. While CoT often yields marginal or scenario-specific improvements, it does not consistently improve overall performance (\eg \texttt{GPT-4o}). Among the models, \texttt{GPT-4.5-Preview} and \texttt{Grok-3-beta} shows the most robust gains, particularly in CR-Multi and IS-Implicit, suggesting its stronger alignment with reasoning-based cues. \texttt{DeepSeek-V3}, however, sees limited benefit, possibly due to the quality of generated reasoning paths. These underscore that CoT prompting is not a universally effective for long-term tool use. Improvements are scenario-dependent and closely tied to the model's inherent reasoning capabilities and contextual retrieval mechanisms. 

\section{Conclusion}
In this work, we introduced \ourD, a novel benchmark designed to rigorously assess the long-term robustness of tool-augmented language models (TALMs) in complex, noisy, and realistic interaction scenarios. Our extensive evaluations across 10 state-of-the-art models demonstrate that while existing TALMs excel in short, clean multi-turn dialogues, their performance significantly degrades in long-term, distraction-rich contexts. This performance gap highlights the critical need for dedicated long-term evaluation frameworks and reveals current limitations in models’ ability to retain context, handle goal shifts, and avoid hallucination over time. \ourD offers a composable, scalable, and diagnostically rich test suite to drive progress toward more reliable, agentic LLM systems. We hope our benchmark facilitates future research aimed at closing the gap between controlled evaluations and the complex realities of real-world tool use.

\section*{Limitations} 
\paragraph{Analysis on the Reasoning-Intensive Models.}
Recently, large reasoning models (LRMs), which are trained to improve reasoning through the use of extended rationales, have demonstrated strong performance across a range of tasks. However, it remains unclear whether such reasoning capabilities directly translate to effective tool use in long-term interactions, which often require pragmatic reasoning over extended contexts. While LRMs may excel at structured reasoning within a single context window, adapting to evolving goals and maintaining coherence across multiple tool-use steps poses additional challenges. Investigating this connection is a promising direction for future work.

\paragraph{Evaluation on Extremely Long-term Interactions.}
In real-world applications, TALMs may interact with users over extended periods, ranging from months to years. To maintain evaluation quality and manageability, our current setup limits the number of sessions. Nevertheless, a valuable future direction is to scale our datasets to hundreds of sessions, enabling evaluation of \textit{in-the-wild} usage patterns and better assessing TALMs' ability to maintain consistency, memory, and utility over long-term interactions.

\paragraph{Number of Used APIs.}
Although our dataset includes a limited number of APIs, our aim is to focus on those that are properly documented, functional, and realistic for use in real-world applications. During the exploration phase, we observed that incorporating APIs that did not meet these criteria significantly degraded data quality. While one could manually construct additional high-quality APIs to expand \ourD, such efforts are beyond the scope of this work.

\section*{Acknowledgments}
This work was supported by Institute of Information \& Communications Technology Planning \& Evaluation (IITP) grant funded by the Korean government (MSIT) (No.RS-2020-II201361, Artificial Intelligence Graduate School Program (Yonsei University)), (No. RS-2024-00457882, National AI Research Lab Project), (2022-0-00077, RS-2022-II220077,AI Technology Development for Commonsense Extraction, Reasoning, and Inference from Heterogeneous Data). Jinyoung Yeo is the corresponding author.

\bibliography{main}
\clearpage

\appendix
\section*{Appendix}
\section{Detailed Related Work}
\label{app:detailed_related_work}
\subsection{Tool-Augmented Language Models}

Large Language Models (LLMs) augmented with external tools have emerged as a promising paradigm for overcoming inherent limitations in static parametric memory~\citep{komeili2022internet} and enabling real-time interaction with the external world~\citep{yang2024codewand, chae2025web}.
The term \emph{tool} encompasses a broad range of functionalities, from simple calculators and Python interpreters~\citep{cai2024latm} to complex API and function calls~\citep{liu2024toolace}. In this work, we focus specifically on the latter\textemdash{}LLMs interfacing with external APIs or invoking function calls to perform tasks.

Early efforts concentrated on enabling LLMs to call tools via prompting and reasoning frameworks. For instance, \citet{parisi2022talm} introduce the concept of Tool Augmented Language Models (TALM), demonstrating that an LLM can iteratively interact with non-differentiable tools through textual interfaces. Similarly, the ReAct framework~\citep{yao2022react} combined chain-of-thought reasoning with action directives, allowing an LLM to plan steps and invoke tools (\eg web search or calculators) in a conversational loop.
Subsequent research has shown that fine-tuning LLMs on tool-use data can significantly enhance their efficacy. Toolformer \citep{schick2023toolformer}, for example, is a fine-tuned model with self-generated tool API calls inserted into its training data, enabling it to decide when and how to use calculators, search engines, translators, and other APIs. 

With the rising demand for practical task-solving capabilities, recent proprietary LLMs have evolved to support tool or function calling as a core interface. Starting with GPT-4's function calling API~\citep{openai_function_calling}, OpenAI has progressively enhanced tool interoperability in their models, culminating in GPT-4.1 and GPT-4o, which natively support multimodal input, extended context windows, and robust tool invocation within conversation flows. Similarly, models like Claude 3.5~\citep{anthropic_claude_3} and Gemini 1.5~\citep{team2024gemini} offer built-in support for external API access and tool orchestration, enabling seamless integration of search, calculator, code execution, and retrieval systems in their agent pipelines. These developments mark a paradigm shift from pure text generation to real-world interaction, where tool use becomes a first-class citizen in the model's interface.
Alongside model-level enhancements, ecosystem-level tool agents have also proliferated. For example, OpenAI's ChatGPT integrates plugin systems for third-party APIs and file browsing, while Google's Bard/Gemini and Anthropic's Claude provide tool-use scaffolding through memory and workflow APIs. 
These systems often combine planning, execution, and memory modules, highlighting the emerging need for robust, API-centric LLM benchmarks and evaluation methods.

Recognizing the performance gap between open-source LLMs and proprietary models in tool calling, recent efforts have focused on equipping smaller or open LLMs with competitive tool-use abilities. For instance, Gorilla~\citep{patil2023gorilla} fine-tunes LLaMA models on a large corpus of API documentation and call traces, leveraging a retriever-generator architecture to improve accuracy and reduce hallucination. ToolLLM~\citep{qin2023toolllm} adopts a reasoning-guided planning strategy that explores multiple API invocation paths via a depth-first search over possible tool sequences. Salesforce's xLAM~\citep{zhang2024xlam} provides a family of open LLMs specifically tuned for tool usage, achieving strong performance on multi-tool benchmarks. Other open initiatives such as ToolAlpaca~\citep{tang2023toolalpaca}, AnyTool~\citep{anytool}, and ToolBridge~\citep{jin2024toolbridge} introduce simulated training pipelines or meta-learning objectives for improving tool generalization in smaller models. These efforts aim to bridge the gap in real-world utility and provide a competitive open-source alternative to commercial tool-augmented LLM systems.

 \subsection{Benchmarks for Evaluating TALMs}

As TALMs have advanced, various benchmarks have been introduced to evaluate their capabilities in using external APIs. Early benchmarks primarily assessed models in static, single-turn settings without environmental feedback. For instance, API-Bank~\citep{li2023apibank} provides 314 task scenarios across 73 APIs, measuring planning, selection, and calling accuracy. Similarly, APIBench~\citep{patil2023gorilla} evaluates whether a model can generate syntactically and semantically correct API calls over a large corpus of public functions. These benchmarks test the model's symbolic reasoning and API schema understanding but lack interactive components.

Subsequent work introduces real-world interaction into the evaluation loop, enabling models to be assessed in dynamic environments. RestGPT \citep{song2023restgpt} connects LLM agents to actual RESTful APIs and evaluates their ability to plan API sequences and recover from execution errors. To further approximate user-facing scenarios, recent benchmarks have introduced user simulators and multi-turn dialogues. ToolDial~\citep{shimtooldial} includes over 11k dialogues annotated with 16 action types, capturing realistic tool-assisted assistant behavior with clarifications, failures, and API chaining. $\tau$-bench~\citep{yao2025taubench} extends this idea by simulating domain-constrained interactions (\eg shopping, airline booking) and measuring goal completion in a transactional database, along with a consistency metric (pass@$k$) to evaluate reliability across repeated trials.

Recent benchmarks have introduced more complex setups to test advanced capabilities like planning and tool use. UltraTool~\citep{ultratool} separates planning from execution to assess models' ability to devise tool-use plans. ToolSandbox~\citep{lu2024toolsandbox} adds \emph{stateful tools} and on-policy user simulators, supporting persistent states and multi-tool chaining. HammerBench~\citep{wang2024hammerbench} expands on this with diverse multi-turn tasks, including Q\&A and argument shifts. ComplexFuncBench~\citep{zhong2025complexfuncbench} addresses the challenge of extremely long contexts (\eg 128k tokens) and proposes constructing long function call sequences.

Despite these developments, a key missing piece in existing benchmarks is support for \textbf{multi-session} evaluation—tracking user preferences, goals, and tool interactions across sessions that reflect the long-term interaction between users and TALMs. Our work addresses this gap by introducing a benchmark that explicitly evaluates TALMs in multi-session settings, enabling more realistic measurement of memory, personalization, and tool-use generalization over time.

\section{Dataset Details}
\begin{table}[t!]
    \centering
    \small
    \renewcommand{\arraystretch}{1.1}
    \setlength{\tabcolsep}{6pt}
    \begin{tabular}{r l r r}
        \toprule
        \textbf{Rank} & \textbf{Category} & \textbf{\# of APIs} & \textbf{Percentage} \\
        \midrule
        1 & Data & 1522 & 11.48\% \\
        2 & Social & 926 & 6.98\% \\
        3 & eCommerce & 845 & 6.37\% \\
        4 & Other & 838 & 6.32\% \\
        5 & Finance & 732 & 5.52\% \\
        6 & Business & 715 & 5.39\% \\
        7 & Entertainment & 690 & 5.20\% \\
        8 & Sports & 677 & 5.11\% \\
        9 & Music & 514 & 3.88\% \\
        10 & Business\_Software & 423 & 3.19\% \\
        11 & Gaming & 397 & 2.99\% \\
        12 & Location & 369 & 2.78\% \\
        13 & Education & 361 & 2.72\% \\
        14 & Travel & 361 & 2.72\% \\
        15 & Email & 356 & 2.68\% \\
        16 & Communication & 301 & 2.27\% \\
        17 & Database & 299 & 2.25\% \\
        18 & Financial & 290 & 2.19\% \\
        19 & Tools & 287 & 2.16\% \\
        20 & Commerce & 287 & 2.16\% \\
        21 & News\_Media & 246 & 1.86\% \\
        22 & Weather & 236 & 1.78\% \\
        23 & Food & 229 & 1.73\% \\
        24 & Media & 218 & 1.64\% \\
        25 & Movies & 141 & 1.06\% \\
        26 & Search & 125 & 0.94\% \\
        27 & Health\_and\_Fitness & 124 & 0.94\% \\
        28 & Science & 113 & 0.85\% \\
        29 & Text\_Analysis & 95 & 0.72\% \\
        30 & SMS & 92 & 0.69\% \\
        \bottomrule
    \end{tabular}
    \caption{Top-30 API categories with number of APIs and percentages.}
    \label{tab:top30_categories}
\end{table}

\subsection{Tool Set}
To construct realistic tool-calling conversations, we collect real-world APIs from RapidAPI from various domains such as travel, finance, and messaging. We provide the statistics of the APIs used for our dataset construction in Table~\ref{tab:top30_categories}. As shown in the table, we utilize various APIs across multiple domains. 

\begin{table*}[t!]
\centering
\resizebox{0.8\textwidth}{!}{%
\begin{tabular}{lccc|ccc|cc}
\toprule
\textbf{Metric} 
& \multicolumn{3}{c|}{\textbf{Context Recall}} 
& \multicolumn{3}{c|}{\textbf{Information Shift }} 
& \multicolumn{2}{c}{\textbf{Missing Context}} \\
\cmidrule(lr){2-4} \cmidrule(lr){5-7} \cmidrule(lr){8-9}
& \textit{Single} & \textit{Multi} 
& & \textit{Explicit} & \textit{Implicit} 
& & \textit{Easy} & \textit{Hard} \\
\midrule
\# Examples            & 93        & 52      &        & 60        & 84        &        & 79        & 85        \\
Utterance Tokens Mean & 1440.71   & 1450.15  &        & 1554.83   & 1542.93   &        & 878.70    & 1339.56   \\
API Tokens Mean       & 6913.02   & 8677.83  &        & 7798.22   & 10797.85  &        & 6372.62   & 9014.56   \\
Total Tokens Mean     & 8353.73  & 10127.98  &        & 9353.05   & 12340.77  &        & 7251.32   & 10354.13  \\
Sessions Mean         & 16.08     & 13.33    &        & 14.22     & 13.58     &        & 11.30     & 12.76     \\
Turns Mean            & 76.77     & 69.94    &        & 76.15     & 78.80     &        & 50.18     & 64.52     \\
\bottomrule
\end{tabular}%
}
\caption{Statistics of \ourD}
\label{tab:dataset_stats}
\end{table*}



\subsection{Dataset Statistics}
\begin{figure*}
    \centering
    \includegraphics[width=0.99\linewidth]{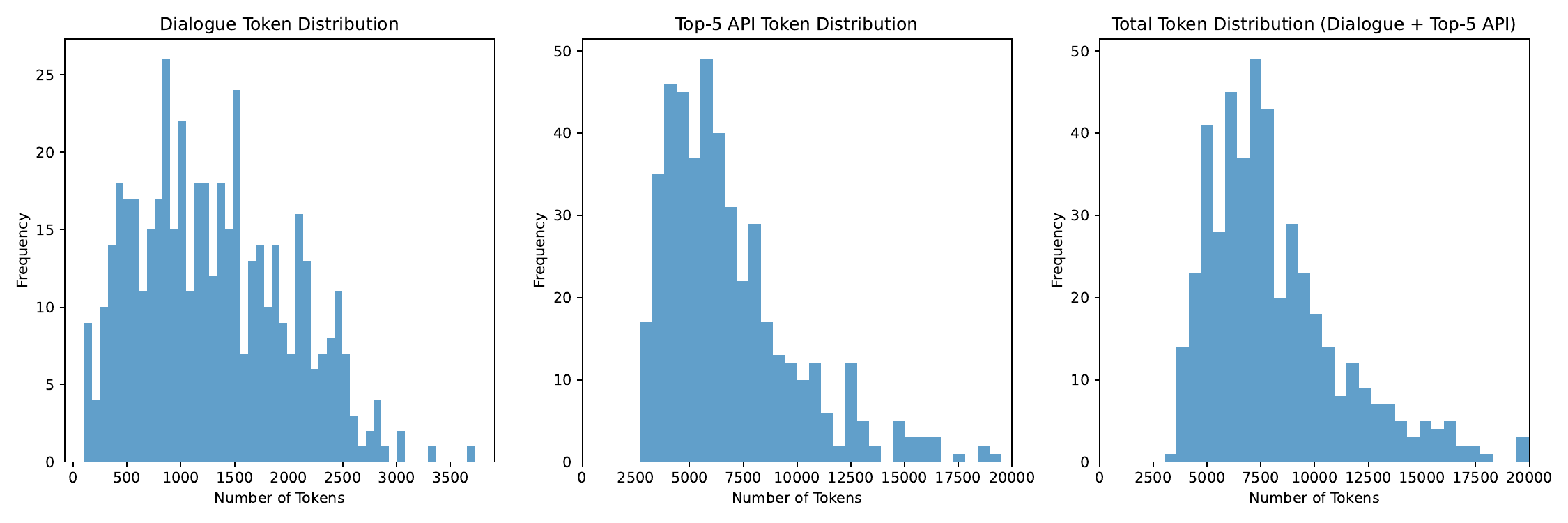}
    \caption{Token length distribution of \ourD}
    \label{fig:token_dist}
\end{figure*}

The statistical overview of \ourD is provided in Table~\ref{tab:dataset_stats}. The dataset comprises three main categories, each split into easy and hard settings. For every category, we include over 100 examples that have been strictly vetted through our filtering pipeline. As demonstrated in Figure~\ref{fig:token_dist}, \ourD features a wide range of token distributions, allowing for robust testing of long-term interaction performance in existing LLMs.

\section{Details in \ourD Construction}
\label{app:dataset_generation_detail}


\subsection{Tool Collection}
We collect over 13K real-world APIs from RapidAPI\footnote{\url{https://rapidapi.com/}}, across domains such as travel, finance, and messaging. APIs lacking stable endpoints or adequate documentation are filtered out. For each selected API, we extract method descriptions, parameter lists, and usage examples. These are normalized into a consistent format using GPT-4o-mini to support downstream tool planning and prompt generation.

\begin{table*}[t!]
    \centering
    \small
    \renewcommand{\arraystretch}{1.2}
    \begin{tabular}{|p{3cm}|p{11cm}|}
        \hline
        \textbf{API Name} & \textbf{Scenario Description} \\
        \hline
        \multirow{3}{*}{\raggedright Search Plant By ID} 
        & A landscape architect from Brazil moves to Canada for a project focusing on sustainable urban gardens. While researching native flora, she needs to identify a specific plant by its ID to ensure compatibility with the local ecosystem. \\
        \cline{2-2}
        & An aspiring botanist in the UK volunteers at a community garden, where he discovers a rare plant species. Curious about its properties and care requirements, he looks it up using the API to provide accurate information to the garden’s committee. \\
        \cline{2-2}
        & A retired school teacher in Australia starts a blog on native plants after moving from the USA. She wants to showcase a particular plant she used to teach about, so she searches for its ID through the API to gather detailed information. \\
        \hline
    \end{tabular}
    \caption{Example scenarios for the \texttt{Search Plant By ID} API}
    \label{tab:search_plant_by_id}
\end{table*}

\subsection{Modeling Tool Sequences}
At the core of each instance lies a tool sequence—a directed list of API calls encoding long-term dependencies. The final target call reuses arguments or outputs from prior needle calls. For example, a hotel booking API might require location and check-in date, both previously produced by a flight search or calendar lookup.

We begin by selecting a target API and then identify prior APIs whose outputs semantically align with its required arguments. This alignment is computed using embedding-based similarity (e.g., Sentence-BERT), followed by LLM-based filtering to remove misleading matches such as unrelated IDs. For each target argument, we annotate whether its source is a previous tool call, user utterance, or remains unresolved (used later in missing context scenarios).

Each sequence is grounded in a plausible situation (e.g., vacation planning, delivery tracking) generated by prompting LLMs to create natural task flows. This provides contextual motivation for tool usage and ensures that dependencies are meaningful and realistic.

For situation generation, we prompt gpt-4o-mini model to generate 30 plausible scenarios related to the API information to gather diverse situation and for each sequence, we randomly sample a single situation from 30 scenarios. We provide generated situation examples for a random API in Table~\ref{tab:search_plant_by_id}.

\subsection{Scenario-Grounded Tool Sequence Construction}
To introduce specific interaction challenges, we embed each tool sequence into one of the \textsc{ToolHaystack} scenarios. For instance, CR scenarios scatter key arguments across earlier dialogue turns; IS scenarios modify goals mid-way through the interaction; MC scenarios mask required arguments to test the model's ability to ask for clarification. These transformations are applied by prompting LLMs to rewrite the original dialogue into one that matches the structural requirements of the scenario, while preserving functional correctness of the tool calls. 

Based on the scenario, we first select the proper APIs that will be included in the tool sequence. For example, for ContextRecall scenario, we choose the API pair with similar parameter(s) as input from the previously aligned API set. Here, for  ContextRecall-\textit{Multi} scenario, we make the API triplet where two APIs share a single argument with the goal API. 

\subsection{Modeling Haystack}
To simulate long-term, noisy conversations, we interleave the tool-relevant dialogue with distractor sessions—unrelated sub-dialogues involving different user intents or casual exchanges. These distractors are independently generated and inserted before, between, or after the tool-relevant content. The number, type, and semantic distance of these distractors are controlled to tune the difficulty of the instance. The result is a single, contiguous dialogue in which critical information is sparsely and irregularly distributed, mimicking the cluttered nature of real-world interactions.

For modeling haystack, we provide previous generated tool sequence as input and randomly select the APIs that can be utilized as haystack. Here, as haystack should not affect the evaluation, we filter out APIs by checking if the randomly selected APIs have any argument that has similar functionality with any argument in the APIs in tool sequence. 

\subsection{Interaction Simulation}
\paragraph{Agent-User Interaction}
Using the tool sequence and scenario specification, we simulate a complete user–agent conversation. The user progressively reveals goals and arguments, while the agent responds through tool calls or natural replies. Each turn is aligned with a structured plan, preserving traceability between user inputs, system actions, and tool responses.

\subsection{Dataset Generation Details}


For ContextRecall-$multi$ scenario, we find that if APIs used for needle include multiple parameters that work similarly with any parameter in the target API, it can repeatedly appear in the dialogue context and makes it easier to find the needle. To this reason, we utilize API triplets where APIs used for needle have a single shared argument with the target API.

For InformationShift scenario, we find that if multiple arguments shifts over turns, most of models fails to generate the correct API call. To this reason, in this scenario, only a single argument evolves in multi-turn conversations.

For MissingContext-$Hard$ scenario, the user query mentions the argument that is needed in detail to prevent the model being confused with other distractor arguments. For example, if the user asked to use OpenAI API in previous session with API key and asks to use Anthropic API without mentioning API key, the model can be confused to use openAI API key. For this situation, we prompt the model to generate detailed query like ``Please use the same API key that I used for Anthropic API previously'' instead of asking ``Please use the same API key that I used''.

\label{sec:dataset_validation}



\section{Evaluation Details}
We use greedy decoding (\ie temperature set to 0) and limit the maximum number of generated tokens to 2,000. For closed-source models, we access the GPT family via the OpenAI API and the Claude models via the Anthropic API. Other models accessible through OpenRouter are used via that service. Finetuned TALMs are served using vLLM~\citep{kwon2023vllm}.

Each model requires specific input formatting; for example, some expect tool descriptions to be provided in JSON rather than Markdown. We follow the formatting guidelines provided in each model's official documentation. All experiments are conducted using three NVIDIA A100 80GB GPUs. All evaluations are conducted in a single run.

\begin{table}[t!]
    \centering
    \resizebox{\textwidth}{!}{%
        \begin{tabular}{lcccccc}
        \toprule
        \textbf{Model} & 
        \makecell{\textbf{Communication}\\\textbf{\& Social}} &
        \makecell{\textbf{Finance}\\\textbf{\& Business}} &
        \makecell{\textbf{Lifestyle}\\\textbf{\& Services}} &
        \makecell{\textbf{Media}\\\textbf{\& Entertainment}} &
        \makecell{\textbf{Utility}\\\textbf{\& Tools}} &
        \textbf{Other} \\
        \midrule
        GPT-4.5-Preview & 0.6667 & 0.3731 & 0.3077 & 0.5465 & 0.4105 & 0.4384 \\
        Claude-3-Opus   & 0.2051 & 0.1493 & 0.0769 & 0.1279 & 0.1579 & 0.1370 \\
        DeepSeek-V3     & 0.2564 & 0.2239 & 0.2308 & 0.3837 & 0.2421 & 0.3562 \\
        \bottomrule
        \end{tabular}%
    }
    \caption{Domain-wise evaluation of model performance.}
    \vspace{-5pt}
    \label{tab:accuracy_by_domain}
\end{table}
\section{Additional Results}
We also conducted a domain-wise evaluation of model performance using a taxonomy that clusters fine-grained API categories into six broader domains. We grouped over 30 fine-grained API categories provided by Rapid API into six high-level domains based on their functional similarity. We report the accuracy of three representative models across the six major domains in the Table \ref{tab:accuracy_by_domain}. These studies demonstrate that ToolHaystack enables controlled, interpretable error analysis across multiple dimensions.





\clearpage
\begin{figure*}[htbp]
  \centering
  \begin{minipage}{\textwidth}
    \begin{tinycodebox}{Prompt for Dialogue Generation (CR-S)}{0.99\textwidth}
## Task Description

You are a skilled dialogue simulator tasked with generating a realistic, multi-session interaction between a human user and a helpful AI assistant equipped with tool-augmented capabilities. Your goal is to produce natural-sounding conversations that reflect a user's evolving needs and information over time, while testing the assistant's long-term memory for multiple arguments across sessions.

You will be given a sequence of sessions involving:

- Evidence session: where the user explicitly provides values for certain arguments.
- Goal session: where the user implicitly refers to earlier information, and the assistant must recall and apply it when using the goal API. The user and the assistant MUST NOT MENTION the parameter values in the goal session.
Each session consists of multiple turns of dialogue between the user and assistant, simulating realistic usage of APIs via natural requests. The assistant should recognize when an API call is needed and formulate the correct parameters using either current or remembered information.

Also, you will be given a scenario for the goal session generation.

## Instructions

1. Carefully read the given information on API.
2. Write the evidence sessions where the user is asking the assistant to use the target API and the assistant asks the user for required parameters.
3. Write the goal session where the user asks the assistant to use the target API and the same shared parameter value that was used in the first session. The user and the AI must not explicitly state the value.
4. Maintain a natural conversation flow while ensuring the assistant collects all required parameters from the user. Note that the conversation should not include numbered lists or item lists.
5. The user should not use vague expressions such as "some data" or "a certain platform". All user requests must be concrete and specific.
6. The user must not mention the API name directly. Instead, the user should describe their goal or the functionality they are trying to achieve. Write the user requests based on the API description.
7. Note that the assistant should be aware the required parameters and ask the user to provide information before the user provides it. The user is not aware of required parameters. 
8. In the goal session, the user should not refer to any previous choices or sessions. It is the assistant's responsibility to retain and reuse the appropriate information from the first session to fulfill the current request.
9. In the goal session, if a shared parameter (e.g., API key) was previously revealed in an evidence session, the user must only say something like "I want to use the one I used before,'' without stating the value. The assistant must not restate the value either but still use it when calling the API.
10. Follow the Output format.

## Input Format

You will be given a structured JSON containing:

- `axis`: always `"ContextRecall-S"`
- `shared_arguments`: the list of arguments that appear across evidence and goal sessions
- `argument_values`: the realistic values associated with those arguments
- `sessions`: a list of session objects, where each object contains:
    - `type`: either `"evidence"` or `"goal"`
    - `api`: the API name used in the session
    - `api_id`: the API id used in the session
    - `arguments`: a dictionary of parameters that should be used in that API call

## Output Format

You must generate one multi-turn session per object in `sessions`. The Output Format should be as below:
[
{{
"session_type": "evidence",
"api_name": "<evidence API name>",
"api_id": "<evidence API id>",
"turns": [
{{ "role": "user", "content": "<natural utterance with request>" }},
{{ "role": "assistant", "content": "<assistant asks for argument value>" }},
{{ "role": "user", "content": "<natural utterance with argument>" }},
{{ "role": "assistant", "content": "<assistant asks for another argument value>" }},

```
... (pairs for every remaining parameter, if any) ...

{{ "role": "user", "content": "<natural utterance with exactly argument N>" }},
{{ "role": "assistant", "content": "Calling the *<API name>* API with {{<argument 1>: <value 1>, <argument 2>: <value 2> ... <argument N>: <value N>}}" }},

```

]
}},
    \end{tinycodebox}
  \end{minipage}
\end{figure*}

\clearpage
\vspace*{0pt}
\noindent\begin{minipage}{\textwidth}
    \begin{tinycodebox}{}{0.99\textwidth}
{{
"session_type": "goal",
"api_name": "<goal API name>",
"api_id": "<goal API id>",
"turns": [
{{ "role": "user", "content": "<natural utterance with request>" }},
{{ "role": "assistant", "content": "<assistant asks for argument value which is not in shared arguments>" }},
... (pairs for every remaining parameter, if any) ...
{{ "role": "assistant", "content": "<assistant asks for shared argument value>" }},
{{ "role": "user", "content": "<high-level request that IMPLIES earlier arguments>" }},
{{ "role": "assistant", "content": "<assistant recalls all values, explicitly states full parameter set, and then executes the API("Calling the *<API name>* API with {{<argument 1>: <value 1>, <argument 2>: <value 2> ... <argument N>: <value N>}}). If some parameter's informations are not given in the evidence sessions, assistant must ask to the user.>" }}
]
}}
]
## Reminder

- If the "session_type" is "evidence", the user should always **explicitly** state the parameters.
- If the "session_type" is "goal", the user and the assistant must not state the parameter values.
- You must note that information between the evidence session and the goal session are shared, so the user must not state the parameter values, but implicitly mention the value by referring to the previous sessions (e.g., I want to use the API key that I used before).

## Input
Scenario:
{scenario}

Sequence:
{multi_session_sequence_json}
  \end{tinycodebox}
\end{minipage}

\begin{figure*}[htbp]
  \centering
  \begin{minipage}{\textwidth}
    \begin{tinycodebox}{Prompt for Dialogue Generation (CR-M)}{0.99\textwidth}
## Task Description

You are a skilled dialogue simulator tasked with generating a realistic, multi-session interaction between a human user and a helpful AI assistant equipped with tool-augmented capabilities. Your goal is to produce natural-sounding conversations that reflect a user's evolving needs and information over time, while testing the assistant's long-term memory for multiple arguments across sessions.

You will be given a sequence of sessions involving:

- Evidence sessions: where the user explicitly provides values for certain arguments.
- Goal session: where the user implicitly refers to earlier information, and the assistant must recall and apply it when using the goal API. The user and the assistant MUST NOT MENTION the parameter values in the goal session.
Each session consists of multiple turns of dialogue between the user and assistant, simulating realistic usage of APIs via natural requests. The assistant should recognize when an API call is needed and formulate the correct parameters using either current or remembered information.

Also, you will be given a scenario for the goal session generation.

## Instructions

1. Carefully read the given information on API and the given scenario.
2. Write the evidence sessions where the user is asking the assistant to use the target API and the assistant asks the user for required parameters. Note that the number of generated evidence sessions should be the same with the number of evidence sessions of the input sequence.
3. Write the goal session where the user asks the assistant to use the target API and the same shared parameter value that was used in the previous session. The user and the AI must not explicitly state the value.
4. Maintain a natural conversation flow while ensuring the assistant collects all required parameters from the user. Note that the conversation should not include numbered lists or item lists.
5. The user should not use vague expressions such as "some data" or "a certain platform". All user requests must be concrete and specific.
6. The user must not mention the API name directly. Instead, the user should describe their goal or the functionality they are trying to achieve. Write the user requests based on the API description.
7. Note that the assistant should be aware the required parameters and ask the user to provide information before the user provides it. The user is not aware of required parameters. 
8. In the goal session, the user should refer to any previous choices or sessions while avoiding mentioning the value of the shared arguments. It is the assistant's responsibility to retain and reuse the appropriate information from the first session to fulfill the current request.
9. In the goal session, as shared parameters (e.g., API key) were previously revealed in evidence sessions, the user must only say something like "I want to use the one I used before,'' without stating the value. The assistant must not restate the value either but still use it when calling the API.
10. Follow the Output format.
11. Very important: In every output session object, the api_name and api_id fields must match exactly the api and api_id values provided for that same session in the input sequence. Do not alter, rename, or re-index them (e.g., do not change RecipeSearch to RecipeFinderAPI, and do not invent new IDs).

## Input Format

You will be given a structured JSON containing:

- `axis`: always `"ContextRecall-M"`
- `shared_arguments`: the list of arguments that appear across evidence and goal sessions
- `argument_values`: the realistic values associated with those arguments
- `sessions`: a list of session objects, where each object contains:
    - `type`: either `"evidence"` or `"goal"`
    - `api`: the API name used in the session
    - `api_id`: the API id used in the session
    - `arguments`: a dictionary of parameters that should be used in that API call

## Output Format

You must generate one multi-turn session per object in `sessions`. The Output Format should be as below:
[
{{
"session_type": "evidence",
"api_name": "<exact evidence API name>",
"api_id": "<exact evidence API id>",
"turns": [
{{ "role": "user", "content": "<natural utterance with request>" }},
{{ "role": "assistant", "content": "<assistant asks for argument value>" }},
{{ "role": "user", "content": "<natural utterance with argument>" }},
{{ "role": "assistant", "content": "<assistant asks for another argument value>" }},

```
... (pairs for every remaining parameter, if any) ...

{{ "role": "user", "content": "<natural utterance with exactly argument N>" }},
{{ "role": "assistant", "content": "Calling the *<API name>* API with {{"<argument 1>": "<value 1>", "<argument 2>": "<value 2>" ... "<argument N>": "<value N>"}}" }},

```

]
}},


    \end{tinycodebox}
  \end{minipage}
\end{figure*}

\clearpage
\vspace*{0pt}
\noindent\begin{minipage}{\textwidth}
    \begin{tinycodebox}{}{0.99\textwidth}
... (additional evidence sessions) ...,
{{
"session_type": "goal",
"api_name": "<exact goal API name>",
"api_id": "<exact gdoal API id>",
"turns": [
{{ "role": "user", "content": "<natural utterance with request>" }},
{{ "role": "assistant", "content": "<assistant asks for argument value which is not in shared arguments>" }},
{{ "role": "user", "content": "<natural utterance with argument>" }},
{{ "role": "assistant", "content": "<assistant asks for another argument value which is not in shared arguments>" }},
... (pairs for every remaining parameter, if any) ...
{{ "role": "assistant", "content": "<assistant asks for shared argument value>" }},
{{ "role": "user", "content": "<high-level request that IMPLIES earlier arguments>" }},
... (pairs for every remaining shared arguments, if any) ...,
{{ "role": "assistant", "content": "<assistant recalls all values, explicitly states full parameter set, and then executes the API("Calling the *<API name>* API with {{"<argument 1>": "<value 1>", "<argument 2>": "<value 2>" ... "<argument N>": "<value N>"}}). If some parameter's informations are not given in the evidence sessions, assistant must ask to the user.>" }}
]
}}
]

## Reminder

- If the "session_type" is "evidence", the user should always **explicitly** state the parameters.
- If the "session_type" is "goal", the user and the assistant must not state the parameter values.
- You must note that information between the evidence session and the goal session are shared, so the user must not state the parameter values, but implicitly mention the value by referring to the previous sessions (e.g., I want to use the API key that I used before).

## Input
Scenario:
{scenario}

Sequence:
{multi_session_sequence_json}
    \end{tinycodebox}
\end{minipage}

\begin{figure*}[htbp]
  \centering
  \begin{minipage}{\textwidth}
    \begin{tinycodebox}{Prompt for Dialogue Generation (IS-E)}{0.99\textwidth}
## Task Description

You are a skilled dialogue simulator tasked with generating a realistic, multi-session interaction between a human user and a helpful AI assistant equipped with tool-augmented capabilities. Your goal is to produce natural-sounding conversations that reflect a user's evolving information over time --- such as updates to addresses, phone numbers, or credentials --- while testing whether the assistant correctly applies the **most recently provided** values during the final session.

You will be given a sequence of sessions involving:
- **Evidence sessions**: where the user provides or updates values for specific arguments (e.g., address, phone number, email). For the shared argument, the user mentions the value with other arguments' values.
- **Goal session**: where the user requests an API operation that requires the assistant to correctly apply the **most recently provided value** of the relevant argument, **without restating it**.

Also, you will be given a scenario for the goal session generation.

The assistant must track user-provided updates across sessions and apply the **latest** value when making an API call in the goal session. This tests the assistant's ability to detect and adopt explicitly stated updates.

## Instructions

1. Carefully read the given information on the APIs and their required arguments.
2. Write evidence sessions where the user provides values, including at least one session where a previously mentioned value is updated without exlicitly mentioning the update (e.g., Instead of "Here is my new email", user should say "You can reach me at XX").
3. In the **goal session**, the user must make a natural request that implies use of the **latest** argument value, but **must not repeat or rephrase the value**.
4. The assistant must infer the updated value from the earlier session(s) and apply it correctly when making the API call while not mentioning the value of the updated value.
5. Maintain a natural dialogue flow (avoid itemized lists or unnatural structure).
6. User requests must be specific, and must not mention the API name directly. Write the user requests based on the API description.
7. Note that the assistant should be aware the required parameters and ask the user to provide information before the user provides it. The user is not aware of what are the required parameters. 
8. The **goal session must refer to past sessions** (e.g., avoid phrases like "as I mentioned earlier" or "same as before").
9. The value of the shared argument MUST NOT be stated in the goal session.
10. Use realistic values for all arguments (e.g., valid addresses, emails, IDs, etc.).
11. "Shared argument" must only be provided in evidence sessions, and never restated in the goal session.

## Output Format

You must generate one multi-turn session per object in `sessions`, formatted as below:

[
{{
  "session_type": "evidence",
  "api_name": "<evidence API name>",
  "api_id": "<evidence API id>",
  "turns": [
    {{ "role": "user", "content": "<natural utterance with request>" }},
    {{ "role": "assistant", "content": "<assistant asks for argument value>" }},
    {{ "role": "user", "content": "<natural utterance with argument>" }},
    {{ "role": "assistant", "content": "<assistant asks for another argument value>" }},
    ...
    {{ "role": "assistant", "content": "Calling the *<API name>* API with {{<argument>: <value>, ...}}" }}
  ]
}},
  {{
    "session_type": "goal",
    "api_name": "<goal API name>",
    "api_id": "<goal API id>",
    "turns": [
      {{ "role": "user", "content": "<natural request (without repeating shared argument)>" }},

      // Assistant asks for each required argument (except shared argument)
      {{ "role": "assistant", "content": "<Asks for arg1>" }},
      {{ "role": "user", "content": "<provides arg1 value>" }},
      {{ "role": "assistant", "content": "<Asks for arg2>" }},
      {{ "role": "user", "content": "<provides arg2 value>" }},
      ...

      // Assistant asks for the shared argument last
      {{ "role": "assistant", "content": "<Asks for shared argument>" }},
      {{ "role": "user", "content": "<please use the latest I gave you>" }},

      {{ "role": "assistant", "content": "Calling the *<API name>* API with {{<arg1>: ..., <arg2>: ..., <shared_arg>: <latest_value>}}" }}
    ]
  }}
]
    \end{tinycodebox}
  \end{minipage}
\end{figure*}

\clearpage
\vspace*{0pt}
\noindent\begin{minipage}{\textwidth}
    \begin{tinycodebox}{}{0.99\textwidth}
## Reminder

- At least one evidence session must update the argument with a new value.
- In the **goal session**, the user must **not repeat or state the updated value**; the assistant must apply the most recent value automatically and must not state it.
- The API call in the **goal session must include the updated value**, not an older one.

## Input
Scenario:
{scenario}

Sequence:
{multi_session_sequence_json}
    \end{tinycodebox}
  \end{minipage}

\begin{figure*}[htbp]
  \centering
  \begin{minipage}{\textwidth}
    \begin{tinycodebox}{Prompt for Dialogue Generation (IS-I)}{0.99\textwidth}
## Task Description

You are a skilled dialogue simulator tasked with generating a realistic, multi-session interaction between a human user and a helpful AI assistant equipped with tool-augmented capabilities. Your goal is to produce natural-sounding conversations that reflect a user's evolving information over time --- such as updates to addresses, phone numbers, or credentials --- while testing whether the assistant correctly applies the **most recently provided** values during the final session.

You will be given a sequence of sessions involving:
- **Evidence sessions**: where the user explicitly provides or updates values for specific arguments (e.g., address, phone number, email).
- **Goal session**: where the user requests an API operation that requires the assistant to correctly apply the **most recently provided value** of the relevant argument, **without restating it**.

Also, you will be given a scenario for the goal session generation.

The assistant must track user-provided updates across sessions and apply the **latest** value when making an API call in the goal session. This tests the assistant's ability to detect and adopt explicitly stated updates.

## Instructions

1. Carefully read the given information on the APIs and their required arguments.
2. Write evidence sessions where the user **explicitly provides values**, including at least one session where a previously mentioned value is **explicitly updated** (e.g., "Here's my new email").
3. In the **goal session**, the user must make a natural request that implies use of the **latest** argument value, but **must not repeat or rephrase the value**.
4. The assistant must infer the updated value from the earlier session(s) and apply it correctly when making the API call while not mentioning the value of the updated value.
5. Maintain a natural dialogue flow (avoid itemized lists or unnatural structure).
6. User requests must be specific, and must not mention the API name directly. Write the user requests based on the API description.
7. Note that the assistant should be aware the required parameters and ask the user to provide information before the user provides it. The user is not aware of required parameters. 
8. The **goal session must refer to past sessions** (e.g., avoid phrases like "as I mentioned earlier" or "same as before").
9. The value of the shared argument MUST NOT be stated in the goal session.
10. Use realistic values for all arguments (e.g., valid addresses, emails, IDs, etc.).
11. "Shared argument" must only be provided in evidence sessions, and never restated in the goal session.

## Output Format

You must generate one multi-turn session per object in `sessions`, formatted as below:

[
{{
  "session_type": "evidence",
  "api_name": "<evidence API name>",
  "api_id": "<evidence API id>",
  "turns": [
    {{ "role": "user", "content": "<natural utterance with request>" }},
    {{ "role": "assistant", "content": "<assistant asks for argument value>" }},
    {{ "role": "user", "content": "<natural utterance with argument>" }},
    {{ "role": "assistant", "content": "<assistant asks for another argument value>" }},
    ...
    {{ "role": "assistant", "content": "Calling the *<API name>* API with {{<argument>: <value>, ...}}" }}
  ]
}},
  {{
    "session_type": "goal",
    "api_name": "<goal API name>",
    "api_id": "<goal API id>",
    "turns": [
      {{ "role": "user", "content": "<natural request (without repeating shared argument)>" }},

      // Assistant asks for each required argument (except shared argument)
      {{ "role": "assistant", "content": "<Asks for arg1>" }},
      {{ "role": "user", "content": "<provides arg1 value>" }},
      {{ "role": "assistant", "content": "<Asks for arg2>" }},
      {{ "role": "user", "content": "<provides arg2 value>" }},
      ...

      // Assistant asks for the shared argument last
      {{ "role": "assistant", "content": "<Asks for shared argument>" }},
      {{ "role": "user", "content": "<please use the latest I gave you>" }},

      {{ "role": "assistant", "content": "Calling the *<API name>* API with {{<arg1>: ..., <arg2>: ..., <shared_arg>: <latest_value>}}" }}
    ]
  }}
]
    \end{tinycodebox}
  \end{minipage}
\end{figure*}
\clearpage
\vspace*{0pt}
\noindent\begin{minipage}{\textwidth}
    \begin{tinycodebox}{}{0.99\textwidth}
## Reminder

- At least one **evidence session must explicitly update** the argument with a new value (e.g., "I moved recently", "Here's my new number").
- In the **goal session**, the user must **not repeat or state the updated value**; the assistant must apply the most recent value automatically and must not state it.
- The API call in the **goal session must include the updated value**, not an older one.

## Input
Scenario:
{scenario}

Sequence:
{multi_session_sequence_json}
    \end{tinycodebox}
  \end{minipage}

\begin{figure*}[htbp]
  \centering
  \begin{minipage}{\textwidth}
    \begin{tinycodebox}{Prompt for Dialogue Generation (MC-E)}{0.99\textwidth}
## Task Description

You are a skilled dialogue simulator tasked with generating a realistic, multi-session interaction between a human user and a helpful AI assistant equipped with tool-augmented capabilities. Your goal is to simulate scenarios where the user provides vague or incomplete references to previously mentioned arguments, and the assistant must carefully determine whether it has sufficient information to proceed with an API call.

Your task is to test the assistant's ability to identify *missing or ambiguous context*---specifically when the user implies a value has already been given, but the assistant actually does **not** have access to a fully specified value in the current or past sessions. The assistant must avoid making incorrect assumptions and should ask for clarification before executing the API.

## Scenario

You are given a sequence of sessions involving:

- Evidence session(s): where the user provides concrete and complete argument values for one or more APIs.
- Goal session: where the user vaguely implies that a certain argument is already known (e.g., "use what I mentioned before"), but the assistant has never actually seen that specific value in the current or any previous session.

Also, you will be given a scenario for the goal session generation.

Each session consists of multiple turns of conversation between the user and assistant. The assistant must detect when a parameter is not clearly specified and ask the user to clarify before making an API call.

## Instructions

1. Use the given API and argument information to build a sequence of natural-sounding dialogues.
2. Write the evidence sessions in which the user provides specific argument values for a certain API.
3. In the goal session, construct a conversation where:
   - For arguments marked as "MISSING", the user should vaguely refer to them as if they were previously mentioned (e.g., "use the page number I mentioned before" or "same page as last time"), even though they were never specified.
   - For all other arguments, the user should explicitly provide these values during the conversation with the assistant.
4. The assistant must recognize that the "MISSING" parameters have not been provided and respond **cautiously**, asking the user to specify these missing arguments.
5. The user in the goal session should not use vague placeholders like "some data" but may use misleading phrases like "as before" or "same as last time" to refer to parameters never actually given.
6. The assistant must **not fabricate values** and should clearly indicate that the argument needs to be specified before proceeding.
7. Note that the assistant should be aware of the required parameters and ask the user to provide information before proceeding with the API call. The user is not aware of required parameters.

## Input Format

You will be given a structured JSON containing:

- `axis`: always `"MissingContext-H"`
- `shared_arguments`: a list of arguments that are implied to be shared but in fact are **not fully specified**
- `argument_values`: values associated with these arguments, provided only in **some sessions** (not necessarily for the goal API)
- `sessions`: a list of session objects, where each object contains:
    - `type`: either `"evidence"` or `"goal"`
    - `api`: the API name used in the session
    - `api_id`: the API id used in the session
    - `arguments`: a dictionary of parameters that should be used in that API call, where "MISSING" indicates arguments the user should vaguely imply were already mentioned

## Output Format

You must generate one multi-turn session per object in `sessions`. The Output Format should be as below:
[
{{
"session_type": "goal",
"api_name": "<goal API name>",
"api_id": "<goal API id>",
"turns": [
{{ "role": "user", "content": "<natural-sounding request that explicitly mentions non-MISSING arguments but vaguely refers to MISSING arguments>" }},
{{ "role": "assistant", "content": "Calling the *Question* API to ask the user about the missing parameter." }}
]
}}
]

## Reminder

- The key evaluation point is whether the assistant correctly identifies when a parameter is **missing or only vaguely implied** and does not proceed without it.
- The user in the **goal session must explicitly provide values for all non-MISSING arguments**, while vaguely implying that MISSING values were previously specified.
- The assistant should request clarification specifically for the MISSING values before calling the API.


## Input
Scenario:
{scenario}

Sequence:
{multi_session_sequence_json}
    \end{tinycodebox}
  \end{minipage}
\end{figure*}

\begin{figure*}[htbp]
  \centering
  \begin{minipage}{\textwidth}
    \begin{tinycodebox}{Prompt for Dialogue Generation (MC-H)}{0.99\textwidth}
## Task Description

You are a skilled dialogue simulator tasked with generating a realistic, multi-session interaction between a human user and a helpful AI assistant equipped with tool-augmented capabilities. Your goal is to simulate scenarios where the user provides vague or incomplete references to previously mentioned arguments, and the assistant must carefully determine whether it has sufficient information to proceed with an API call.

Your task is to test the assistant's ability to identify *missing or ambiguous context*---specifically when the user implies a value has already been given, but the assistant actually does **not** have access to a fully specified value in the current or past sessions. The assistant must avoid making incorrect assumptions and should ask for clarification before executing the API.

## Scenario

You are given a sequence of sessions involving:

- Evidence session(s): where the user provides concrete and complete argument values for one or more APIs.
- Goal session: where the user vaguely implies that a certain argument is already known (e.g., "use what I mentioned before''), but the assistant has never actually seen that specific value in the current or any previous session.

Also, you will be given a scenario for the goal session generation.

Each session consists of multiple turns of conversation between the user and assistant. The assistant must detect when a parameter is not clearly specified and ask the user to clarify before making an API call.

## Instructions

1. Use the given API and argument information to build a sequence of natural-sounding dialogues.
2. Write the evidence sessions in which the user provides specific argument values for a certain API.
3. In the goal session, construct a conversation where the user vaguely implies that one or more arguments were already provided---even though they were not. The assistant must recognize this and respond **cautiously**, asking the user to specify any missing arguments.
4. The user in the goal session should not use vague placeholders like "some data" but may use misleading phrases like "as before" or "same as last time" to refer to parameters never actually given.
5. The assistant must **not fabricate values** and should clearly indicate that the argument needs to be specified before proceeding.
6. Note that the assistant should be aware the required parameters and ask the user to provide information before the user provides it. The user is not aware of required parameters. 


## Input Format

You will be given a structured JSON containing:

- `axis`: always `"MissingContext-H"`
- `shared_arguments`: a list of arguments that are implied to be shared but in fact are **not fully specified**
- `argument_values`: values associated with these arguments, provided only in **some sessions** (not necessarily for the goal API)
- `sessions`: a list of session objects, where each object contains:
    - `type`: either `"evidence"` or `"goal"`
    - `api`: the API name used in the session
    - `api_id`: the API id used in the session
    - `arguments`: a dictionary of parameters that should be used in that API call

## Output Format

You must generate one multi-turn session per object in `sessions`. The Output Format should be as below:
[
{{
  "session_type": "evidence",
  "api_name": "<evidence API name>",
  "api_id": "<evidence API id>",
  "turns": [
    {{ "role": "user", "content": "<natural utterance with request>" }},
    {{ "role": "assistant", "content": "<assistant asks for argument value>" }},
    {{ "role": "user", "content": "<natural utterance with argument>" }},
    {{ "role": "assistant", "content": "<assistant asks for another argument value>" }},
    ...
    {{ "role": "assistant", "content": "Calling the *<API name>* API with {{<argument>: <value>, ...}}" }}
  ]
}},
{{
  "session_type": "goal",
  "api_name": "<goal API name>",
  "api_id": "<goal API id>",
  "turns": [
    {{ "role": "user", "content": "<natural utterance with request>" }},
    {{ "role": "assistant", "content": "<natural response asking for the value of the required argument which is not the shared argument>" }},
    {{ "role": "assistant", "content": "<natural response sharing the value of the required argument which is not the shared argumen>" }},
    ... (pairs for every remaining parameter except the shared argument) ...
    {{ "role": "assistant", "content": "<naturally asking for the shared argument value>" }},
    {{ "role": "user", "content": ""<saying that you already know it as I used it before when I was <explanation about the goal api description, which can be differentiate it with apis used in evidence sessions> without mentioning the value. To distinguish the shared argument from the shared argument utilized before, clearly mention the function of the API.>" }}
    {{ "role": "assistant", "content": "Calling the *<API name>* API with {{<shared argument>: "MISSING", ...}}" }}
  ]
}}
]
    \end{tinycodebox}
  \end{minipage}
\end{figure*}

\begin{figure*}[htbp]
  \centering
  \begin{minipage}{\textwidth}
    \begin{tinycodebox}{}{0.99\textwidth}
## Reminder

- The key evaluation point is whether the assistant correctly identifies when a parameter is **missing** and fill the parameter with "MISSING".
- The user in the **goal session must not give the missing value upfront**, but should imply it's known.
- The last utterance of the user explaning that the shared argument alogn with the explanation about the goal api description MUST explain about the goal API instead of apis used in the evidence sessions. It must be separated.

## Input
Scenario:
{scenario}

Sequence:
{multi_session_sequence_json}
    \end{tinycodebox}
  \end{minipage}
\end{figure*}

\begin{figure*}
\begin{minipage}{0.99\textwidth}
    \begin{smallcodebox}{Prompt for Filtering Dialogue}{0.99\textwidth}
You are a senior evaluator tasked with evaluating a generated tool using conversation.

## Original prompt
{prompt}

## Generated Conversation
{response}

## Constraints
- If the "session_type" is "evidence", the user should always **explicitly** state the parameters.
- If the "session_type" is "goal", the user and the assistant must not state the parameter values.
- You must note that information between the evidence session and the goal session are shared, so the user must not state the parameter values, but implicitly mention the value by referring to the previous sessions (e.g., I want to use the API key that I used before).

## Your Task
- If **any** of the constraints above are not satisfied, output **NO**.
- If **all** constraints are satisfied, output **YES**.
- Output only one word: **YES** or **NO**. Do not add explanations or any other text.


    \end{smallcodebox}
\end{minipage}
\end{figure*}

\begin{figure*}
\begin{minipage}{0.99\textwidth}
    \begin{smallcodebox}{Prompt for Filtering Tool Sequence (CR-S)}{0.99\textwidth}
You are a senior evaluator tasked with evaluating a generated tool-call sequence.

## Original prompt
{prompt}

## Generated Sequence
{response}

## API Call Label
{label}

## Constraints
- Ensure the label includes **all required and optional parameters** defined by the API.

## Your Task
- If **any** of the constraints above are not satisfied, output **NO**.
- If **all** constraints are satisfied, output **YES**.
- Output only one word: **YES** or **NO**. Do not add explanations or any other text.

    \end{smallcodebox}
\end{minipage}
\end{figure*}

\begin{figure*}
\begin{minipage}{0.99\textwidth}
    \begin{smallcodebox}{Prompt for Filtering Tool Sequence (CR-M)}{0.99\textwidth}
You are a senior evaluator tasked with evaluating a generated tool-call sequence.

## Original prompt
{prompt}

## Generated Sequence
{response}

## API Call Label
{label}

## Constraints
- Ensure the label includes **all required and optional parameters** defined by the API.
- If a **shared argument** appears in multiple evidence sessions, the value in the goal session **must match the latest value** from the most recent evidence session.

## Your Task
- If **any** of the constraints above are not satisfied, output **NO**.
- If **all** constraints are satisfied, output **YES**.
- Output only one word: **YES** or **NO**. Do not add explanations or any other text.


    \end{smallcodebox}
\end{minipage}
\end{figure*}

\begin{figure*}
\begin{minipage}{0.99\textwidth}
    \begin{smallcodebox}{Prompt for Filtering Tool Sequence (IS)}{0.99\textwidth}
You are a senior evaluator tasked with evaluating a generated tool-call sequence.

## Original prompt
{prompt}

## Generated Sequence
{response}

## API Call Label
{label}

## Constraints
- Ensure the label includes **all required and optional parameters** defined by the API.

## Your Task
- If **any** of the constraints above are not satisfied, output **NO**.
- If **all** constraints are satisfied, output **YES**.
- Output only one word: **YES** or **NO**. Do not add explanations or any other text.

    \end{smallcodebox}
\end{minipage}
\end{figure*}

\begin{figure*}
\begin{minipage}{0.99\textwidth}
    \begin{smallcodebox}{Prompt for Filtering Tool Sequence (MC)}{0.99\textwidth}
You are a senior evaluator tasked with evaluating a generated tool-call sequence.

## Original prompt
{prompt}

## Generated Sequence
{response}

## API Call Label
{label}

## Constraints
- Ensure the label includes **all required and optional parameters** defined by the API.

## Your Task
- If **any** of the constraints above are not satisfied, output **NO**.
- If **all** constraints are satisfied, output **YES**.
- Output only one word: **YES** or **NO**. Do not add explanations or any other text.

    \end{smallcodebox}
\end{minipage}
\end{figure*}

\begin{figure*}
\begin{minipage}{0.99\textwidth}
    \begin{tinycodebox}{Prompt for Generting Distractor}{0.99\textwidth}
You are an expert conversation designer tasked with generating *distractor session sequences* for evaluating a large language model's ability to ignore irrelevant information in multi-session tool-augmented interactions.

You are given an original multi-session input containing shared argument values and one or more goal-relevant APIs. Your task is to generate realistic, well-structured **distractor evidence sessions** that are functionally and semantically unrelated to the original session content.

These distractor sessions should simulate plausible user queries and assistant tool-use, but must NOT contain any references to shared arguments, overlapping parameters, or semantically similar API goals.

## Constraints:

- All distractor sessions must be labeled with `"type": "distractor"`.
- Each session must follow the same structural format as original sessions.
- Use only APIs from the provided API Subset.
- Selected APIs must NOT:
  - Be present in the original sessions.
  - Have parameters that overlap with any parameters used in the evidence or goal sessions.
  - Serve a similar purpose to any of the original APIs (e.g., if the original APIs involve Amazon product data, avoid ecommerce/product-related APIs).
- Arguments used in distractor sessions must:
  - Be realistic and appropriate for the selected API.
  - NOT match or reuse any values from `shared_arguments`.

## API Selection:
- Choose N diverse APIs from the API subset that are clearly irrelevant to the goal or evidence APIs from the original session.
- Ensure domain and functional diversity (e.g., do not cluster around the same topic or use only financial or health APIs).

## Output Format:
Return a list of N distractor session objects, each in the following format:

```json
{{
  "type": "distractor",
  "api": "<distractor API name>",
  "api_id": "<distractor API id>",
  "api_description": "<distractor API description>",
  "arguments": {{
    "<param_name_1>": "<realistic value>",
    "<param_name_2>": "<realistic value>",
    ...
  }}
}}

### Input

<original_session_data>
{input_sequence}
</original_session_data>

<api_subset>
{api_subset}
</api_subset>

<num_distractor_sessions>
{num_session}
</num_distractor_sessions>

### Task:
Generate <num_distractor_sessions> distractor session objects as defined above, ensuring each one strictly complies with the constraints and format.

Only return the list of session objects in the specified JSON format---do not add extra commentary or explanations.


    \end{tinycodebox}
\end{minipage}
\end{figure*}

\begin{figure*}
\begin{minipage}{0.99\textwidth}
    \begin{smallcodebox}{Prompt for Evaluation}{0.99\textwidth}
Your task is to generate the API call that the AI should call after the given conversation. 

1. Read the conversation.
2. Select the API that will be used for the given conversation among the given API set.
3. Based on the API information, generate the API call.
4. The API call should be in dictionary format where the keys are api_name and parameters.
5. Make sure that the output contains all required parameters and optional parameters.
6. If the user did not mention about the optional parameters, fill the values as their default value. If the default value is not stated, fill the value with empty string ''.
7. The output should not be in code snippet. Generate API call dictionary only.
8. Output only a valid Python dictionary. Do not include any explanations, markdown code blocks, or extra text. The output must start with '{{' and end with '}}'.

## Output format
{{
    "api_name": "<api_name>", # Case sensitive, use the exact api name in the document below
    "parameters": 
        {{
            "<parameter1>": "<value1>",
            "<parameter2>": "<value2>",
            ...
            "<parameter_n>": "<value_n>",
        }}
}}

Here is the conversation and information about the API.

### Conversation
{all_sessions}

### API set
{api_set}

    \end{smallcodebox}
\end{minipage}
\end{figure*}

\end{document}